\pdfoutput=1
\documentclass[conference]{IEEEtran}
\IEEEoverridecommandlockouts
\usepackage{cite}
\usepackage{amsmath,amssymb,amsfonts}
\usepackage{algorithmic}
\usepackage{graphicx}
\usepackage{adjustbox}
\usepackage{subcaption}
\usepackage{caption}
\usepackage[dvipsnames]{xcolor}
\usepackage{tikz}
\usepackage{academicons}
\definecolor{orcidlogocol}{HTML}{A6CE39}
\usepackage{tabularx}
\usepackage[inkscapelatex=false]{svg}
\usepackage{hyperref}
\hypersetup{
    bookmarks=false,
    colorlinks=true,
    citecolor=green,
    linkcolor=blue,
    filecolor=magenta,
    hypertexnames=true,
    urlcolor=cyan,
    pagebackref=true,
    pdftitle={IV2023},
    pdfpagemode=FullScreen
    }
\usepackage[capitalize]{cleveref}
\crefname{section}{Sec.}{Secs.}
\Crefname{section}{Section}{Sections}
\Crefname{table}{Table}{Tables}
\crefname{table}{Tab.}{Tabs.}
\definecolor{lime}{HTML}{A6CE39}
\DeclareRobustCommand{\orcidicon}{%
    \begin{tikzpicture}
    \draw[lime, fill=lime] (0,0)
    circle [radius=0.16]
    node[white] {{\fontfamily{qag}\selectfont \tiny ID}};
    \draw[white, fill=white] (-0.0625,0.095)
    circle [radius=0.007];
    \end{tikzpicture}
    \hspace{-2mm}
}
\newcommand{\orcidWalter}{\href{https://orcid.org/0000-0003-4565-1272}{\orcidicon}}
\newcommand{\orcidJoseph}{\href{https://orcid.org/0000-0003-2263-6491}{\orcidicon}}
\newcommand{\orcidMarcel}{\href{https://orcid.org/0000-0002-8780-5762}{\orcidicon}}
\newcommand{\orcidTung}{\href{https://orcid.org/0000-0002-7908-6112}{\orcidicon}}
\newcommand{\orcidStefan}{\href{https://orcid.org/0000-0001-6799-647X}{\orcidicon}}
\newcommand{\orcidBohan}{\href{https://orcid.org/0000-0002-2948-5529}{\orcidicon}}
\newcommand{\orcidKnoll
}{\href{https://orcid.org/0000-0003-4840-076X}{\orcidicon}}
\usepackage{textcomp}
\usepackage{siunitx}
\def\BibTeX{{\rm B\kern-.05em{\sc i\kern-.025em b}\kern-.08em
    T\kern-.1667em\lower.7ex\hbox{E}\kern-.125emX}}

\makeatletter 
\newcommand{\linebreakand}{%
  \end{@IEEEauthorhalign}
  \hfill\mbox{}\par
  \mbox{}\hfill\begin{@IEEEauthorhalign}
}
\makeatletter
\newcommand*{\emails}[2][@tum.de]{%
    \def\@tempa{\@gobble}%
    \@for\qrr@email:=#2\do{%
        \edef\@tempb{\noexpand\href{mailto:\qrr@email #1}{\qrr@email}}%
        \edef\@tempa{\unexpanded\expandafter{\@tempa}{, }\unexpanded\expandafter{\@tempb}}}%
    \{\@tempa\}#1%
}
\begin{document}
\title{InfraDet3D: Multi-Modal 3D Object Detection based on Roadside Infrastructure Camera and LiDAR Sensors\\
}
\author{
Walter Zimmer\orcidWalter,
\and
Joseph Birkner\orcidJoseph,
\and
Marcel Brucker\orcidMarcel,
\and
Huu Tung Nguyen\orcidTung,
\linebreakand
Stefan Petrovski\orcidStefan,
\and
Bohan Wang\orcidBohan,
\and
Alois C. Knoll\orcidKnoll
\thanks{$^{1}$The authors are with the School of Computation, Information and Technology (CIT), Department of Informatics, Technical University of Munich, TUM, 85748 Garching-Hochbrueck, Germany.\newline Contact:
  {\tt\small walter.zimmer@tum.de}
}%
}
\maketitle
\begin{abstract}
Current multi-modal object detection approaches focus on the vehicle domain and are limited in the perception range and the processing capabilities. Roadside sensor units (RSUs) introduce a new domain for perception systems and leverage altitude to observe traffic. Cameras and LiDARs mounted on gantry bridges increase the perception range and produce a full digital twin of the traffic. In this work, we introduce \textit{InfraDet3D}, a multi-modal 3D object detector for roadside infrastructure sensors. We fuse two LiDARs using early fusion and further incorporate detections from monocular cameras to increase the robustness and to detect small objects. Our monocular 3D detection module uses HD maps to ground object yaw hypotheses, improving the final perception results. The perception framework is deployed on a real-world intersection that is part of the \textit{A9 Test Stretch} in Munich, Germany. We perform several ablation studies and experiments and show that fusing two LiDARs with two cameras leads to an improvement of $\boldsymbol{+1.90}$ \texttt{mAP} compared to a camera-only solution. We evaluate our results on the A9 infrastructure dataset and achieve $\boldsymbol{68.48}$ \texttt{mAP} on the test set. The dataset and code will be available at \url{https://a9-dataset.com} to allow the research community to further improve the perception results and make autonomous driving safer.
\end{abstract}
\begin{IEEEkeywords}
3D Perception, Camera-LiDAR Fusion, Roadside Sensors, Infrastructure Sensors, Autonomous Driving
\end{IEEEkeywords}
\begin{figure}[htbp]
\centering
\minipage{0.49\linewidth}
  \includegraphics[width=\linewidth]{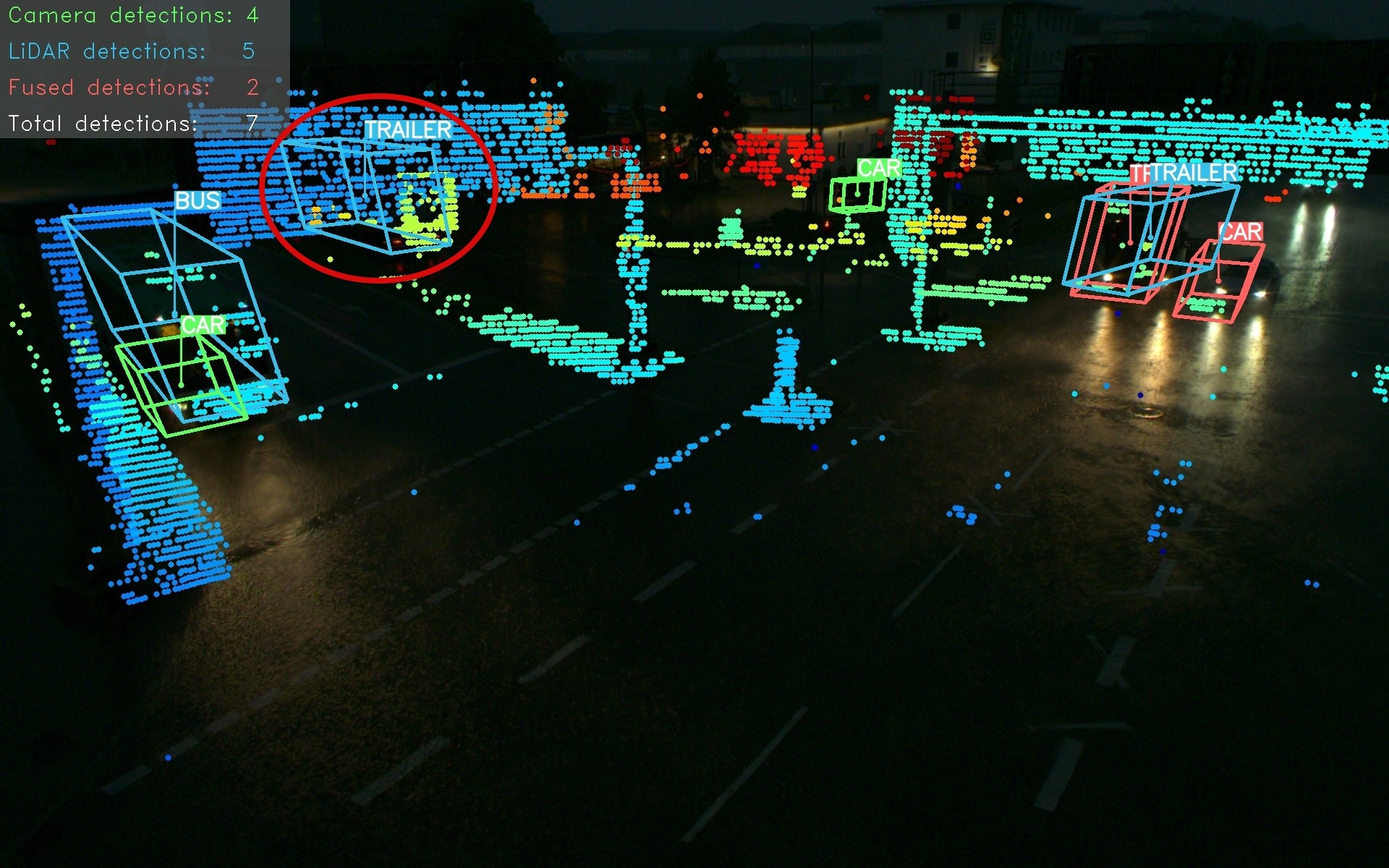}
\endminipage
\minipage{0.49\linewidth}
  \includegraphics[width=\linewidth]{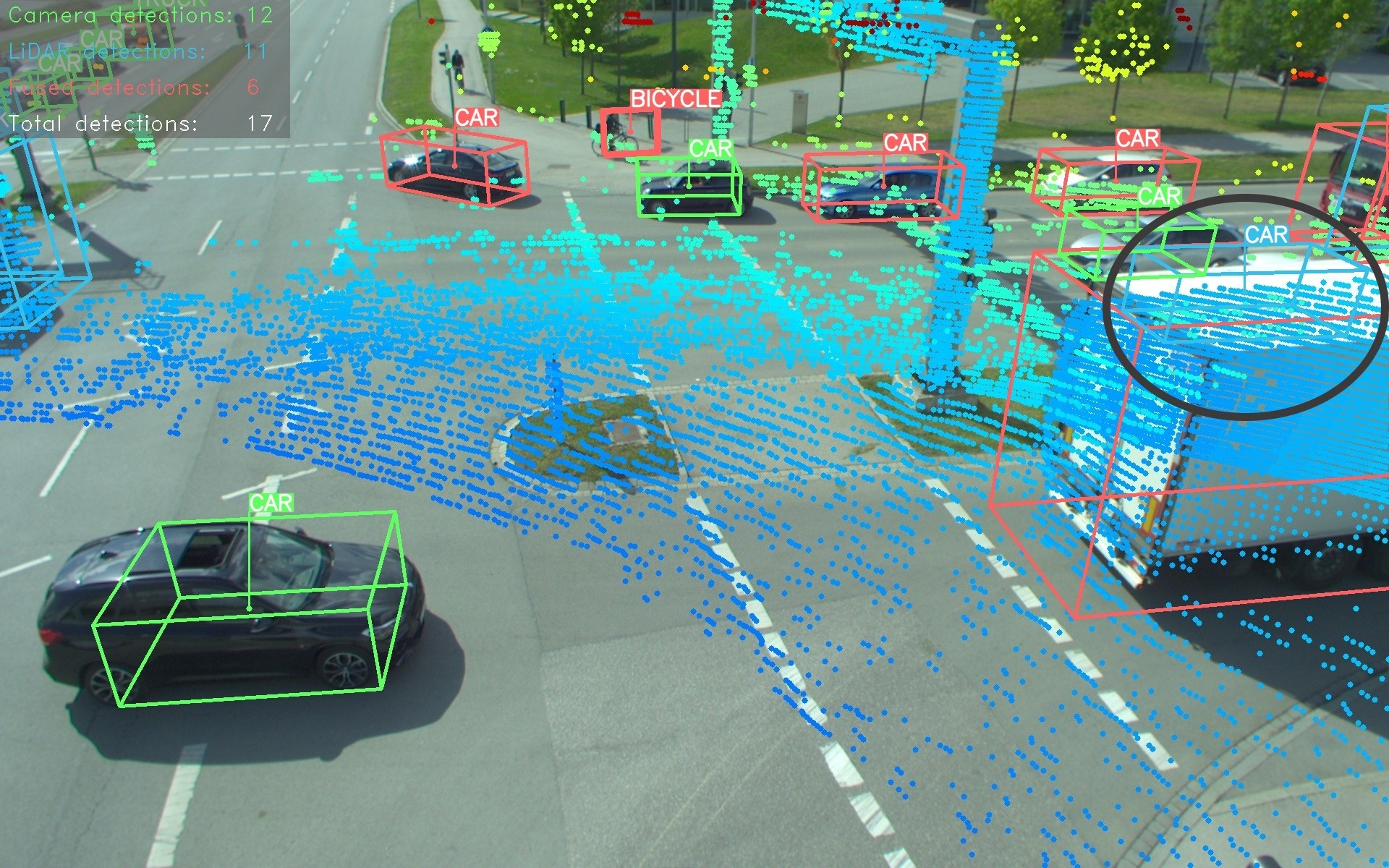}
\endminipage\hfill
\minipage{0.49\linewidth}
  \includegraphics[width=\linewidth,trim={0cm 7.94cm 0 7.94cm},clip]{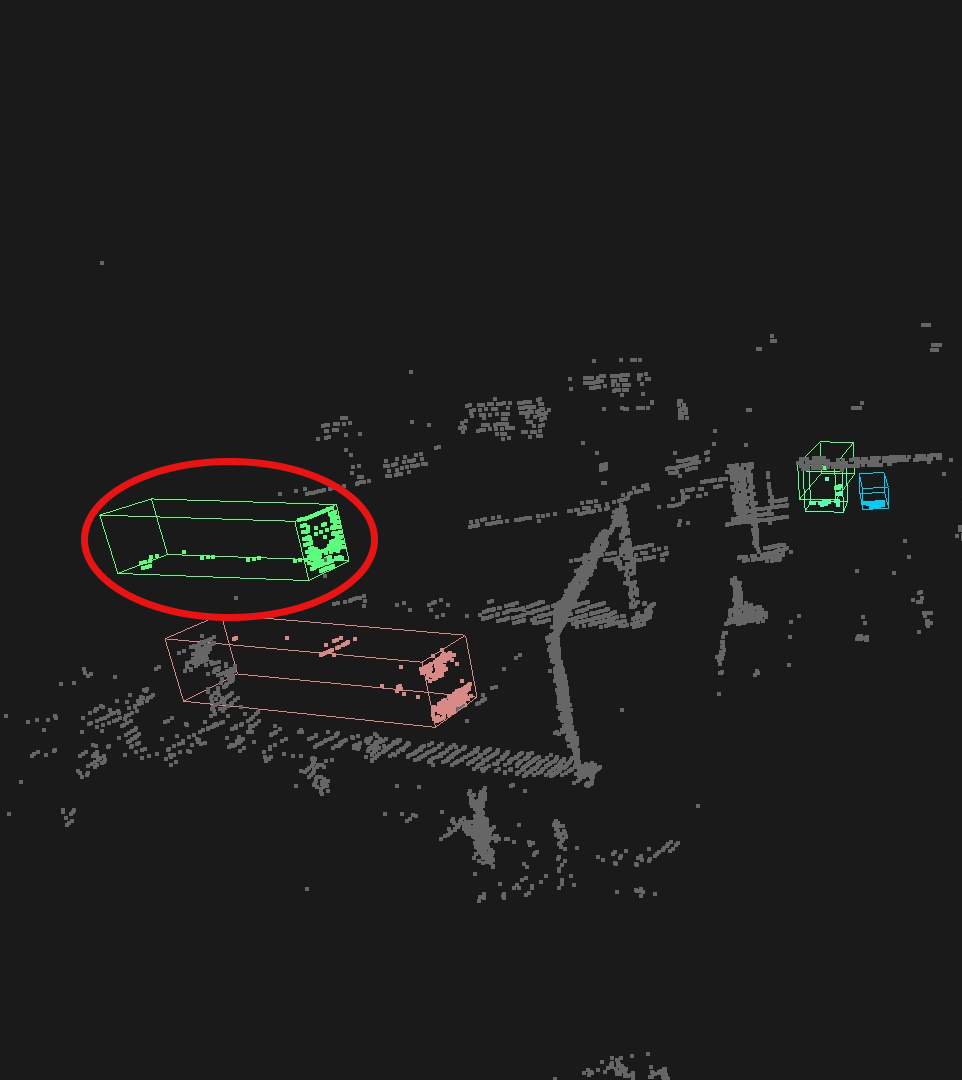}
\endminipage
\minipage{0.49\linewidth}
  \includegraphics[width=\linewidth,trim={0cm 9cm 0 9cm},clip]{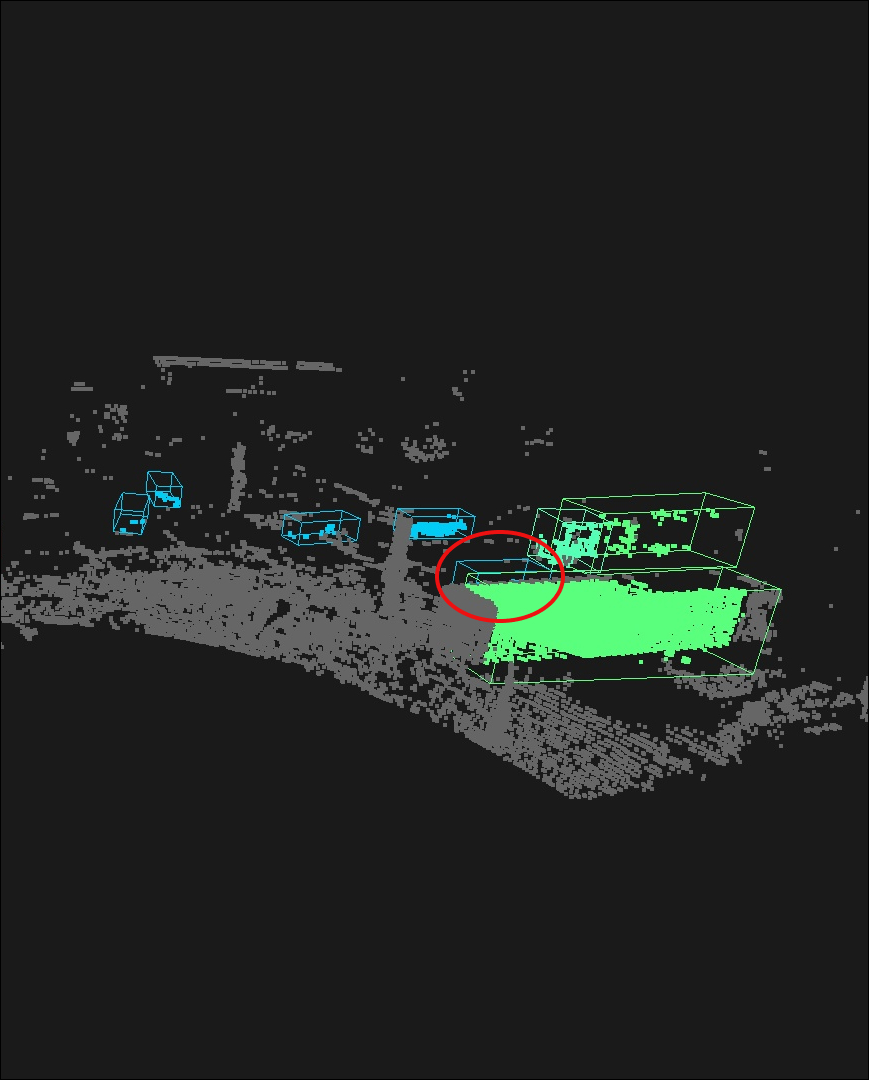}
\endminipage
\caption{Early and late fusion of two roadside cameras and LiDARs. We register point clouds from two LiDARs using G-ICP \cite{zhou2016fast} and project them with the camera-LiDAR detections into the image. \textit{Left column:} Night detection results in more and better classified LiDAR detections. \textit{Right column:} Detections during day time  demonstrate a 41.67\% increase in detections using the fusion approach. Moreover, even occluded objects, like the car behind the trailer (right) or the truck behind the gantry bridge (left), can be detected with our \textit{InfraDet3D} Fusion Framework.}
\label{fig:main_figure}
\end{figure}

\section{Introduction}
Roadside perception is vital to improve the situation awareness and to provide a far-reaching view for automated vehicles. Roadside sensors installed on infrastructure systems like the A9 Test Stretch \cite{krammer2022providentia,cress2021intelligent} increase the perception range drastically. They perceive objects around the corner, e.g. to warn drivers performing a left or right turn. A cost-effective solution is needed to process perception models in real-time and provide accurate results at the same time.

\begin{figure*}[t]
    \centering
    \includegraphics[clip, trim=0cm 6cm 0cm 0cm, width=1.00\textwidth]{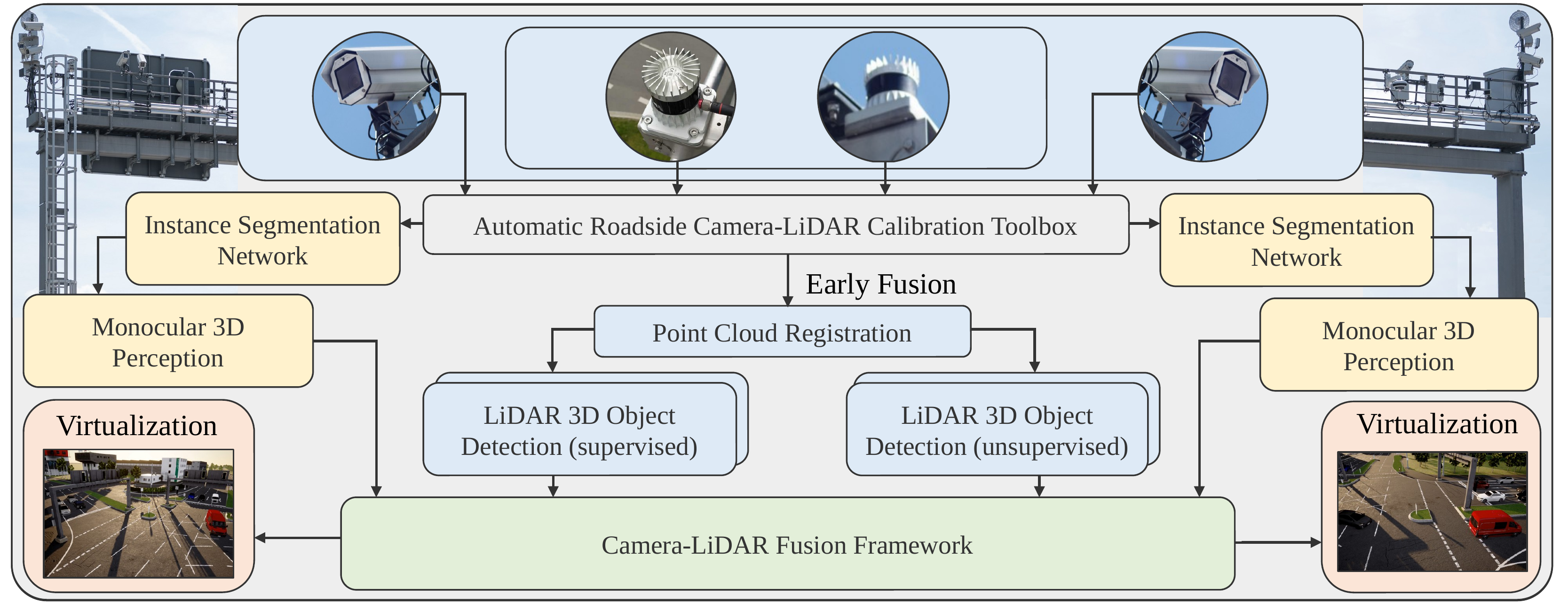}
    \caption{\textit{InfraDet3D} Perception Framework Architecture. Our proposed model is deployed on a real intersection (S110) part of the A9 Test Stretch for Autonomous Driving in Munich, Germany.}
    \label{fig:my_label}
\end{figure*}

Positional data captured from roadside sensors is sent through high performance units to all traffic participants to decrease blind spots and prevent accidents. It has been shown that roadside sensors increase the situation awareness by sending important notifications and warnings to vulnerable road users (VRUs) and drivers \cite{zimmer2022survey,zimmer2022real,zimmer2022realdomain}. In this work, we contribute to the challenge of sparse point clouds in the domain of roadside perception in the following way:
\begin{itemize}
    \item We propose a real-time point cloud registration algorithm to register infrastructure LiDARs which enhances the point density. Our experiments show that early fusion of point clouds leads to an increase of $+1.32$ \texttt{mAP}.
    \item Fusing supervised and unsupervised LiDAR 3D object detectors increases the robustness and reduces the number of false positive detections.
    \item We connect our perception module to real HD maps (+2.7 \texttt{mAP}) of the \textit{A9 Testbed} to extract road information, as well as to validate and filter the perception results.
    \item Our camera-LiDAR fusion module further enhances the robustness of our whole perception toolbox (+1.62 \texttt{mAP}) by providing perception results during day and night time.
    \item Finally, we evaluate all 3D detectors on the A9-I dataset and introduce a leaderboard to allow the research community to benchmark their models on our dataset.
\end{itemize}

\section{Related Work}
Much research has been done in the area of roadside 3D perception. Traditional approaches \cite{gong2021pedestrian} increase the robustness of roadside LiDAR perception systems because of the similarity and the lack of diversity in the background point cloud. Furthermore, they do not require labeled data and process point clouds efficiently. In \cite{guo2021detection} a 3D vehicle detection approach is proposed that uses a single camera. First, they segment the instance mask in the image, extract the bottom contour and project it on the road plane to get the 3D position. Then, they cluster the projected points into objects by applying K-means clustering. Afterwards, they estimate the dimensions (length and width) and orientation (heading angle) of vehicles by fitting a box for each cluster. Finally, they refine the 3D box to fit it within the 2D box by maximizing the posterior probability.
Bai et al. proposes a learning-based approach \cite{bai2022cyber} that requires huge labeled datasets and performs poorly in domains where no labeled data is available. The authors introduce a real-time LiDAR-based traffic surveillance system to detect objects in 3D. They develop \textit{3DSORT}, a 3D multi-object tracker by extending \textit{DeepSORT} \cite{wojke2017simple}. The limitation of all mentioned approaches is that they have no labeled training data of roadside LiDARs and use open-source datasets like \textit{nuScenes} \cite{caesar2020nuscenes} to train the model.
To the best of our knowledge there is no roadside 3D perception framework available that is able to fuse data from multiple road side sensor units. Furthermore, there is no solution that combines different fusion levels (early and late fusion), as well as traditional and learning-based approaches into a single framework.

\section{A9 Intersection Dataset}
The A9 Intersection (A9-I) dataset is an extension of the A9 Dataset \cite{cress2022a9}. It contains labeled data (in \textit{OpenLABEL} format) of two cameras and two LiDAR sensors mounted on the S110 gantry bridge that is part of the \textit{A9 Test Stretch for Autonomous Driving}. It contains 9,600 labeled point clouds and images with 57,743 labeled 3D objects ($\varnothing$12/frame) and is split into a training (80\%), validation (10\%), and test set (10\%). The test set contains a sequence with labeled track IDs and sampled frames from four different scenarios. We applied stratified sampling to balance the dataset among sensor types and scenarios. The set contains 25\% night data with severe weather conditions like heavy rain which allows the model to perform well under challenging weather conditions. Our dataset was created by labeling experts and some improvements were done to further enhance the label quality using the \textit{proAnno} labeling toolbox which is based on \cite{zimmer20193d}.

\section{Sensor Calibration}
\begin{figure}[htbp]
	\centering
	\includegraphics[width=1.0\linewidth]{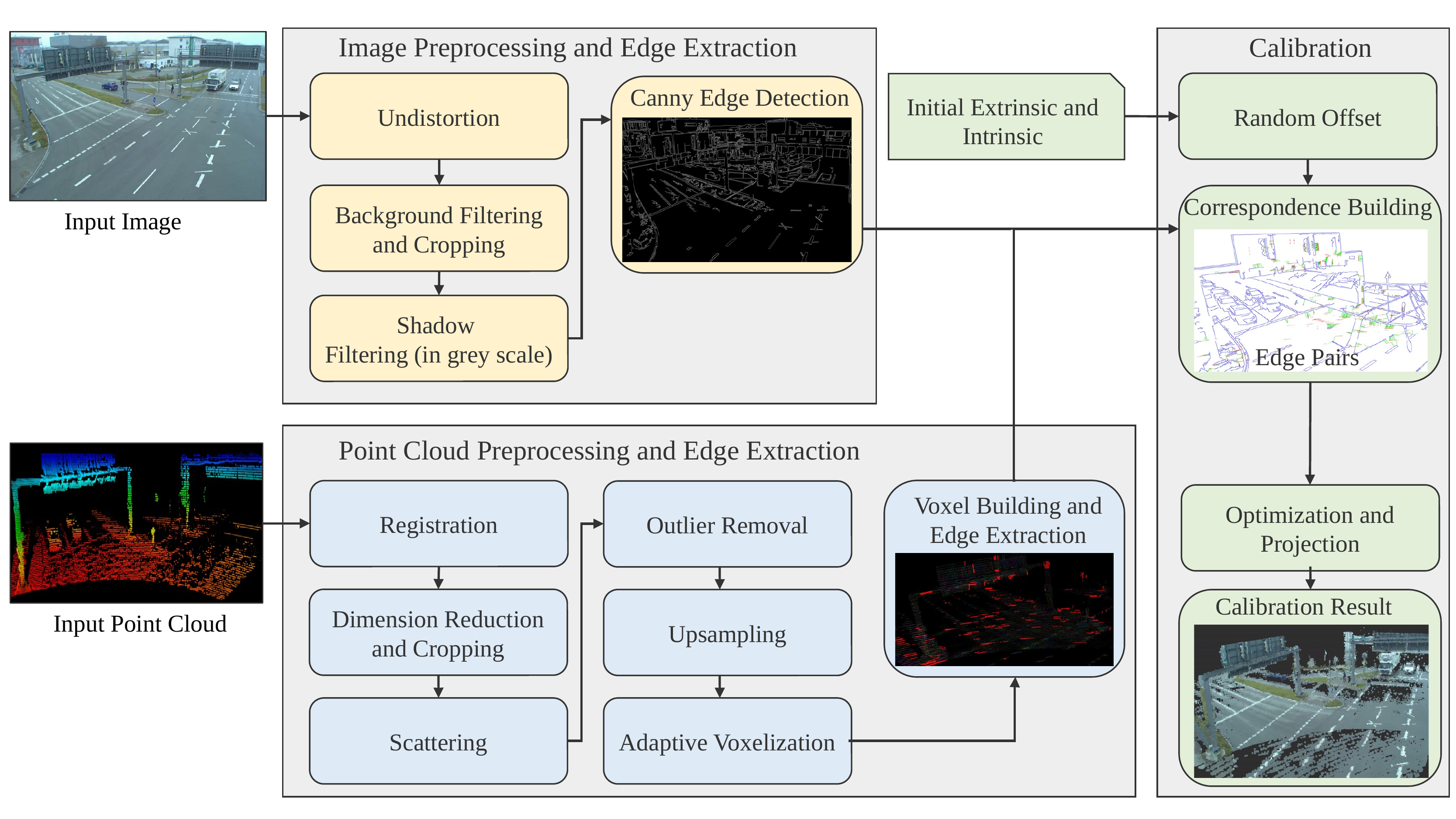}
	\caption{Automatic calibration pipeline. We integrate four camera image and seven LiDAR point cloud preprocessing modules into our pipeline in order to increase the robustness of real-world outdoor calibration of roadside sensors. The algorithm takes the image and point cloud that is published continuously on the live system as input and outputs both, the calibration results and qualitative projections of point clouds into camera images.}
	\label{fig:auto}
\end{figure}

In our framework multiple roadside LiDAR and camera sensors are fused and processed together for the detection task. Our automatic calibration of infrastructure LiDARs and cameras, which outputs the precise pose of these sensors, is the most fundamental part of the framework. In order to calibrate the sensors in the real world, we propose an automatic target-less LiDAR-camera calibration model. We use the calibration method proposed in \cite{pixel_level} as a baseline and extend it to outdoor scenes captured by infrastructure roadside sensors of a different manufacturer. To improve the robustness of the model under different external conditions, such as different scene complexities, lighting conditions, or sensor conditions, we introduce various automatic preprocessing submodules (see \Cref{fig:auto}).\\

First, we undistort the input images. After that, an automatic background cropping (based on monocular depth estimation \cite{MiDaS}) is employed to remove the background objects. If there is shadow on the ground, the automatic shadow filtering module will be activated to filter the shadow. After the preprocessing, the \textit{Canny} edge detector \cite{CANNY} is adopted to extract 2D edges in images.\\
For LiDAR preprocessing, point clouds from three LiDARs are registered to the target LiDAR. The input point cloud is cropped and only four dimensions are preserved (x,y,z and intensity). Scattering is applied to increase the density of single frame point cloud scans. Afterwards, the point cloud is automatically subdivided into ground and non-ground point clouds. Outlier removal is applied to the ground point cloud to filter the noise in order to preserve more points of the gantry bridge. We also use point upsampling \cite{Daniel} to improve the surface texture of point clouds. After the preprocessing, voxels are extracted from the point clouds. For faster extraction, adaptive voxelization \cite{mlcc} is introduced. \textit{RANSAC} plane fitting is applied to extract planes within the voxel. The intersections among planes are extracted as LiDAR edge clouds.\\
After the edges are extracted from the point cloud, they are projected into the image and correspondences between LiDAR and camera edges are established. A cost based on maximum likelihood estimate is optimized and the qualitative result is generated. Our automatic calibration model demonstrates good robustness against different weather conditions and traffic scenarios in the intersection and provides accurate extrinsic calibration values for the perception framework.
\begin{figure*}[t]
    \centering
    \includegraphics[width=\linewidth]{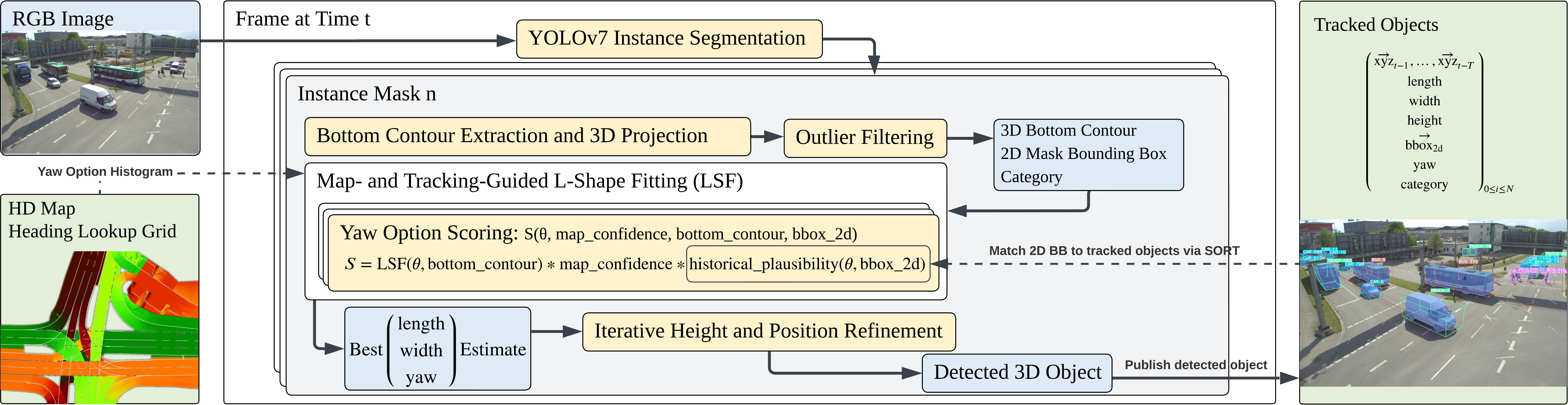}
    \caption{Monocular 3D object detection pipeline, grounding shape hypotheses via tracking and the HD map.}
    \label{fig:mono3d-arch}
\end{figure*}

\section{Monocular 3D Object Detection}
Due to their low cost and high output information density, monocular RGB cameras are incorporated as sensors into the \textit{InfraDet3D} architecture. The monocular detection pipeline is based on an augmented \textit{L-Shape-Fitting} algorithm as proposed by \cite{zhang2017efficient}. The basic \textit{L-Shape-Fitting} algorithm has also been used in other recent roadside infrastructure perception architectures, such as the detector for the MONA dataset \cite{gressenbuch2022mona} and the Cooperative Vehicle Infrastructure System 3D detector \cite{guo2021detection}. However, the augmentation of this algorithm with object tracking, to score yaw hypotheses based on historical plausibility, is novel. Furthermore, we propose the integration of the High-Definition (HD) map to limit yaw hypotheses with regard to matching lanes. Both features are inspired by TrafficNet \cite{rezaei2021traffic} and UrbanNet \cite{carrillo2021urbannet} architectures. An overview of the full monocular detection pipeline is given in Figure \ref{fig:mono3d-arch}.

\subsection{From 2D Instance Masks to 3D Bottom Contours}
We use the YOLOv7 Instance Segmentation model \cite{wang2022yolov7} on RGB camera frames. The RGB frames are downscaled to 1280x720 pixels size, to accelerate the instance segmentation runtime. The instance masks, which are output from the model, are processed to extract the bottom image contour from the masks. The 2D bottom contour coordinates for each mask are then projected from screen-space to 3D intersection space via raycasting. Finally, the \textit{DBSCAN} (Density-Based Spatial Clustering of Applications with Noise) \cite{schubert2017dbscan} algorithm is applied to denoise each detection's 3D bottom outline.

\subsection{HD Map Yaw Candidate Lookup}
Using lane geometry from an HD map of the sensor-covered areas, each lane's road surface is rasterized into a heading lookup grid, covering the field of view of the respective camera. The heading lookup grids are rendered to a resolution of 10x10 cm grid cells. Each grid cell $C_{ij}$ is a set $\{(\text{lane\_id}_k, \theta_k)\}_{k=0}^{k \leq N_{ij}}$ of lane ID and heading pairs which apply to the respective cell. The heading for a lane at the position of the grid cell is interpolated from the direction of the surrounding lane borders. At inference time, for each 3D bottom contour point of a detected object, the grids are queried to compute a set $L=\{(\text{lane\_id}_i, \theta_i)\}_{i=0}^{i \leq N}$ of possible heading values along the bottom contour. This set is aggregated into a histogram with hit counts and average heading angle per lane ID. The hit counts for each lane ID are normalized into confidence values in the range of $[0,1]$ through division over the maximum hit count value. This yields one $H_j=\{(\text{lane\_id}_i, \theta_i, \text{confidence}_i)\}_{i=0}^{i<M}$ three-tuple-set of possible heading values for each instance j.

\subsection{Augmented L-Shape-Fitting}
The \textit{L-Shape-Fitting} (LSF) algorithm searches for a rectangle that fits a specific bottom contour by maximizing a score value\footnote{Such as negative average variance}, which is calculated as a function of a rectangle yaw ($\theta$) hypothesis and the 3D bottom contour points. In the basic form, the algorithm simply goes through several $\theta$ values from the range $[0, \pi]$ at fixed increments. In our augmented version of the algorithm, we only run LSF for $\theta$ values as present in the HD map lookup histogram for each 3D bottom contour. Furthermore, we multiply the calculated score value for each yaw hypothesis with the respective confidence value from the normalized map lookup histogram. Finally, the score is also multiplied with a historical plausibility factor. The calculation of this factor is explained in the following. \\

Using a screen-space SORT tracker \cite{bewley2016simple}, we match a detected object's bounding box to detections from previous frames. For a successfully matched detection, historical 3D position values $L=\{\vec{l}_{t-1},\dots,\vec{l}_{t-T}\}$ are retrieved. Given the historical positions $L$ and a position hypothesis $\vec{l}_t(\theta_t)$, the historical plausibility score $\text{HP}$ for a yaw hypothesis $\theta_t$ is calculated as in the following equation:
$$\text{HP}=\prod_{\delta_t=1}^{\delta_t \leq T}\pi/2 - \Delta_\angle(|\theta_{t} - \text{atan2}(\vec{l}_t(\theta_t)-\vec{l}_{t-\delta_t})| \mathtt{~mod~}\pi)$$

The Delta-Angle function $\Delta_\angle: [0, \pi) \rightarrow [0, \pi/2)$ converts the passed raw angular difference, which is already less than $\pi$, into a value less than $\pi/2$ by returning angular deltas $\delta_{>\pi/2}$ larger than $\pi/2$ as $\pi-\delta_{>\pi/2}$. This ensures that a yaw hypothesis, which is parallel, yet opposed to a historical orientation, is not erroneously punished. In practice, we have implemented a threshold of six historical positions that are evaluated to determine the plausibility of a yaw hypothesis.

\subsection{Height Estimation and Dimension Filtering}
The height for each detection is initialized from a fixed value for the object type of the detection. Both the height and the location are then jointly optimized through binary search, until the estimated projected 2D object height and the original mask height are the same by $\epsilon<1\text{px}$.  The length and width values, as estimated by the \textit{L-Shape-Fitting} algorithm for each 3D bottom contour, are limited to minimum and maximum values, which are also looked up per object category.

\section{LiDAR 3D Object Detection}
\subsection{Unsupervised 3D Object Detection}
LiDAR sensors are a popular choice for roadside object detection as they provide accurate 3D information in a large field of view and are lighting invariant. Studies on roadside LiDAR object detection favor traditional approaches based on clustering. Before clustering an extracted foreground point cloud into individual objects, these studies discard the ground, walls, trees, and other background artifacts from the raw point cloud. To discard the irrelevant background, our first 3D LiDAR object detector uses a fast four step procedure. First, the detector crops a predefined region of interest, which always remains the same as the LiDAR sensor is installed statically on roadside infrastructure. This first step removes 69.9\% of points on average. Second, the detector finds points belonging to the ground by considering the Euclidean distance to a predefined plane model together with a threshold of 0.2~m. Third, the detector filters background artifacts within the region of interest based on the coarse-fine triangle algorithm \cite{Zhang.2022}. The fourth step is radius outlier removal ($n = 15$, $r = 0.8$), which refines the extraction of the foreground point cloud.
The remaining foreground point cloud represents all traffic objects, including stationary ones. It is divided into distinct point clusters, each corresponding to a potential road user, by \textit{DBSCAN} ($\epsilon = 0.8$, $n_{min} = 3$). Around each point cluster, the detector fits an oriented 3D bounding box using its convex hull and principal component analysis. Finally, the detector classifies the localized objects by means of object dimensions and point density.

\subsection{Supervised 3D Object Detection}
For the data-driven approach, we are using \textit{PointPillars} \cite{lang2019pointpillars} which runs with a fast inference rate of $38$ FPS. In comparison to the unsupervised approach, we can input the registered point cloud (262k points) directly into the model, consisting of three modules. In the first step the \textit{PillarFeatureNet} converts the point cloud into a sparse pseudo-image. After obtaining the pseudo-image, the 2D backbone produces features at a small spatial resolution. Theses features are then upsampled and concatenated. In the last step, an anchor-based detection head tries to match the bounding boxes to the ground truth.

We used the \textit{PointPillars} implementation of \textit{OpenPCDet} \cite{openpcdet2020} and adapted it to our A9 intersection dataset. For training, we limited the point cloud range from $-64$ to $64$~m in x-y direction and from $-8$ to 0~m in z direction. In the feature extraction step, we set the voxel size to $[0.16, 0.16, 0.8]$. The model was trained on 10 classes for $160$ epochs and optimized using Adam with a learning rate of $\alpha=0.003$, weight decay of $0.01$ and cyclic momentum of $\beta=0.9$.

\section{Multi-Modal 3D Object Detection}
\begin{figure}[htbp]
    \centering
    \includegraphics[width=1.0\linewidth]{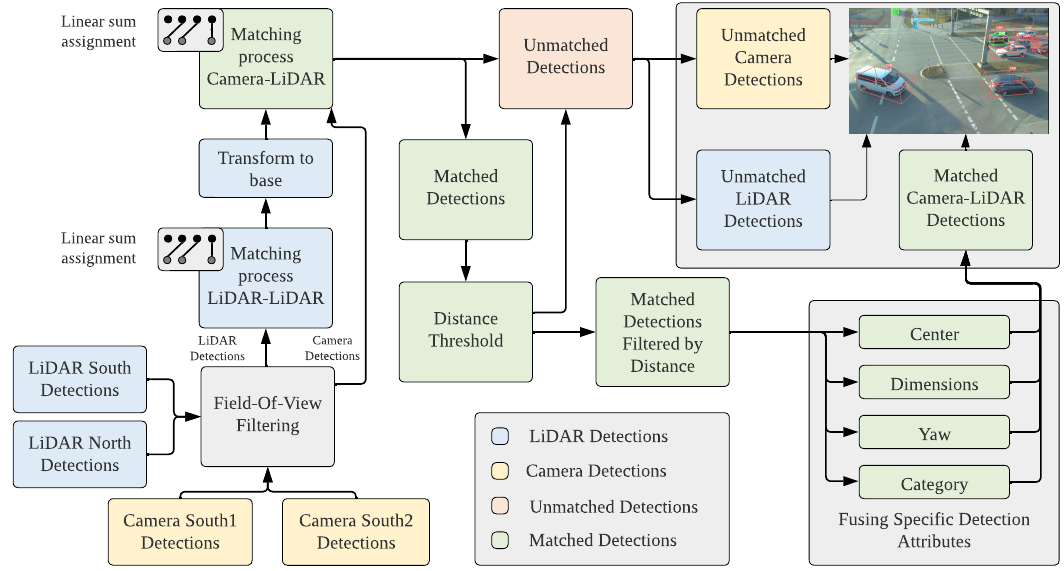}
    \caption{Multi-modal 3D object detection pipeline. We apply a camera field-of-view filtering for all detections.}
    \label{fig:multi_pipeline}
\end{figure}

For the fusion of both modalities (LiDAR and camera detections) a late fusion technique is applied (see Fig. \ref{fig:multi_pipeline}).

\subsection{Data Association}
A widely adopted method for combining and matching sensor data at the later stage is through data association, also defined as the linear assignment problem (LAP). It finds a one-to-one mapping between two sets of elements, such that the sum of the assigned pairwise costs is minimized.

The \textit{Jonker-Volgenant} algorithm \cite{jonker1988shortest} is a method for solving the LAP and is based on augmenting paths. The algorithm starts by finding an initial feasible solution, e.g. by using the \textit{Hungarian} algorithm \cite{kuhn1955hungarian}. Then, it repeatedly searches for an augmenting path, a path of alternating unmatched and matched elements that starts and ends at an unmatched element, and increases the number of assigned elements by one. The algorithm stops when no augmenting path can be found - the solution is optimal.

The \textit{modified Jonker-Volgenant} algorithm \cite{crouse2016implementing} is a variation of the original one that improves its performance by using a heuristic search strategy. The heuristic builds on the idea of prioritizing the search for augmenting paths that are expected to have a high gain in terms of reducing the total cost.

In this work, the \textit{modified Jonker-Volgenant} algorithm is chosen due to its increased speed ($O(n^3)$ \cite{crouse2016implementing}) in comparison to its variants. It also works well with non-integer costs. In our case, the matching process took $0.008~ms$ on average per frame on the test set on a AMD Ryzen 5800X 8-Core CPU with an average number of $14.55$ objects per frame.

\begin{table*}[htbp]
    \centering
    \scalebox{0.89}{
    \begin{tabular}{|l|l|l|l|r|r|r|}
        \hline
        \textbf{Model} & \textbf{Modality} & \textbf{Fusion Level} & \textbf{Dataset} & \textbf{Precision} & \textbf{Recall} & $\boldsymbol{mAP_{3D}}$\\
        \hline
         MonoDet3D (Ours)                         & Image south1             & -            & A9-I south1 & 48.12 & 59.23 & 49.01 \\
                                                  & Image south2             & -            & A9-I south2 & 27.84 & 29.11 & 26.33 \\
                                                  & Image south1+south2             & LF  & A9-I full   & 37.98 & 44.17 & 37.67 \\
         \hline
         LidarDet3D (Ours)                        & Point Cloud S+N      & EF & A9-I full   & 8.40 & 6.32 & 8.13\\
                                                  & Point Cloud S+N      & LF  & A9-I full   & 6.34 & 5.43 & 6.10\\
         \hline
         PointPillars*\cite{lang2019pointpillars} & Point Cloud N   & -            & A9-I south1 & \underline{56.66} & \underline{57.44} & \underline{56.10}\\
                                                  & Point Cloud N   & -            & A9-I south2  & 20.96 & 31.79 & 20.62\\
                                                  & Point Cloud S   & -            & A9-I south2 & \textbf{36.32} & \textbf{48.93} & \textbf{35.75} \\

                                                  & Point Cloud S   & -            & A9-I south1  & 35.37 & 50.01 & 34.81\\
                                                  & Point Cloud S+N & EF & A9-I full    & \textbf{62.85} & \textbf{51.22} & \textbf{62.11} \\
                                                  & Point Cloud S+N & LF  & A9-I full    & \underline{46.97} & \underline{51.23} & \underline{46.10}\\
         \hline
         \textbf{InfraDet3D} (Ours):  & Image south1 + Point Cloud S+N & LF of (Image + Point Cloud EF)  & A9-I south1  & \textbf{68.83} &	\textbf{74.89}	&  \fontsize{6}{6}{\color{ForestGreen}{(\textit{+12.38})}} \fontsize{8}{8}{\textbf{68.48}}\\
                        MonoDet3D + PointPillars           & Image south2 + Point Cloud S+N & LF of (Image + Point Cloud EF)  & A9-I south2  & \underline{33.52} & \underline{44.57} &  \fontsize{6}{6}{\color{red}{(\textit{-2.54})}} \underline{33.21}\\
                                                         & Image (south1+south2) + Point Cloud S+N & LF of (Image LF + Point Cloud LF)   & A9-I full    & 38.93 & 49.94 &  38.58\\
                                                         & Image (south1+south2) + Point Cloud S+N & LF of (Image LF + Point Cloud EF)   & A9-I full    & 39.28 & 48.12 &  38.86\\
         \hline
    \end{tabular}
    }
    \caption{Evaluation results on the A9-I intersection test set (N=North, S=South, EF=Early Fusion, LF=Late Fusion). \\\hspace{\textwidth}We report the $mAP_{3D@0.1}$ results for the following six classes: \textit{Car}, \textit{Truck}, \textit{Bus}, \textit{Motorcycle}, \textit{Pedestrian}, \textit{Bicycle}. * PointPillars inference score threshold is set to 0.3.}
    \label{tbl:evaluation}
\end{table*}

\subsection{Early Fusion of LiDAR Sensors}
Our first fusion module combines multiple point cloud scans from different LiDAR sensors at time step \textit{t} into a single dense point cloud. We preprocess the point clouds, as described in \cite{zimmer2022real}. First, we downsample the point cloud and estimate point normals. Then, we compute a 33-dimensional FPFH\footnote{Fast point feature histogram} feature vector \cite{rusu2009fast} for each point. This feature describes the local geometric property of each 3D point. Afterwards, we register several point clouds from roadside LiDARs that are time-synchronized with an NTP time server. The point cloud registration algorithm makes use of Fast Global Registration \cite{zhou2016fast} to provide an initial transformation. For the refinement of the transformation, we use point-to-point ICP \cite{besl1992method} as it leads to a lower RMSE value (0.448 m) than point-to-plane ICP. The full registration pipeline of two Ouster OS1-64 LiDARs takes 18.36 ms (54 FPS) on an Intel Core i7-9750H CPU and a voxel size of 2~m.

\subsection{Late Fusion of LiDAR Sensors}
For the LiDAR-to-LiDAR late fusion, we operate in LiDAR coordinate space. We transform the detections obtained by the unsupervised LiDAR detector and the supervised LiDAR detector into a common coordinate system.
We match detections based on a distance of 3~m between their central positions. Matched detections are merged by selecting the central position and yaw vector of the detected object from the LiDAR sensor closest to the detection. Dimensions of the merged detections are computed by calculating the mean average of the detections from both detectors. Additionally, all unmatched detections are also included in the final result, resulting in an increase of 12.93\% in the number of detections compared to using only a single LiDAR sensor.

\subsection{Camera-LiDAR Late Fusion}
For the camera-LiDAR fusion, we transform the LiDAR detections into the base coordinate system of the gantry bridge, which serves as the coordinate system for obtaining the monocular detections. This step is crucial for computing the inter-detection distances between camera and LiDAR instances based on their respective center positions. After the linear sum assignment, the matched detections are further filtered by a distance threshold of 3~m. The attributes of the matched detections are merged by eliminating matched camera detections and retaining only matched LiDAR detections, as they demonstrate greater accuracy on average during evaluation. The integration of the HD map leads to a substantial improvement (see Table \ref{tab:mono3d_ablation}) in the camera yaw result, however it remains inferior to the results obtained from LiDAR. Table \ref{tab:detection_improvement} displays the dependence of the \texttt{mAP} increase on the various attributes. \\

\begin{table}[htbp]
\caption{Ablation study for matched camera-LiDAR detections calculated for south1 camera using early and late fusion. Taking LiDAR detection attributes leads to $\texttt{mAP}_{3D}$ score improvements in all cases.}
\begin{center}
\begin{tabular}{|l|c|}
\hline
\textbf{Fused Attribute~~~~~~~~~~~~~~~} & \textbf{~~~~~Improvement in} $\boldsymbol{mAP_{3D}}$~~~~~~\\
\hline
Center position & +2.96 \\
Yaw & +0.16 \\
Dimensions & +1.65 \\
Category & +13.10 \\
\hline
Total improvement & \textbf{+17.87}\\
\hline
\end{tabular}
\label{tab:detection_improvement}
\end{center}
\end{table}

\section{Evaluation}
\subsection{Monocular Perception - L-Shape-Fitting Augmentations}
To determine the impact of the aforementioned augmentations on the quality of the 3D pose estimation, we evaluated the \textit{L-Shape-Fitting} algorithm in several configurations on the categories (\textit{Car}, \textit{Bus}, \textit{Truck}, \textit{Motorcycle}) of the A9 infrastructure dataset.
The ablation study results of the \textit{L-Shape-Fitting} augmentation evaluations are presented in Table \ref{tab:mono3d_ablation}.

\begin{table}[htbp]
\caption{Ablation study of \textit{L-Shape-Fitting} (LSF) augmentations on the vehicle category superset of the A9-I dataset.}
\scalebox{0.92}{
\begin{tabular}{|l|l|l|}
\hline
\textbf{Configuration} & $\boldsymbol{mAP_{3D}}$ & $\boldsymbol{IoU_{3D}}$ \\
\hline
Basic \textit{L-Shape-Fitting} (LSF) & $62.01$ & $0.29$ \\
LSF with HD map yaw confidence & \underline{$64.71$} & \underline{$0.43$} \\
LSF with hist. plausibility via \textit{SORT} tracking & $48.42$ & $0.31$  \\
LSF with both augmentations & $\textbf{67.65}$ & $\textbf{0.44}$ \\
\hline
\end{tabular}
\label{tab:mono3d_ablation}
}
\end{table}

The ablation study confirms that tracking and historical plausibility alone are not useful to improve over basic \textit{L-Shape-Fitting}. With the addition of the HD map, however, the risk that an earlier bad yaw choice propagates into the future is greatly reduced, and the historical plausibility further increases the gain in \texttt{mAP} from $+2.7$ to $+5.64$.

\subsection{Monocular 3D Perception - Performance Considerations}
As presented, the monocular 3D object detection pipeline achieves a throughput of 22 FPS in our test bench setup using an RTX 2080S GPU with 1280x720 24-bit RGB input frames. This is limited by the performance of the \textit{YOLOv7} instance segmentation inference time. At 640x480 resolution, the frame rate increases to 66 FPS using \textit{TensorRT}.

\subsection{LiDAR 3D Perception - Runtime Evaluation}
Our unsupervised 3D detector achieves a processing speed of $47$ FPS as Table \ref{tab:results_runtime} demonstrates.
\begin{table}[htbp]
\caption{Runtime evaluation of detector modules on the A9-I test set. All modules are implemented in Python 3.8 and run on a 2.9 GHz dual-core Intel Core i5 CPU.}
\begin{center}
\begin{tabular}{|l|c|}
\hline
\textbf{Module~~~~~~~~~~~~~~~~~~~~~~~~~~~~~~} & ~~~~~~~\textbf{$\varnothing$ Runtime in} \si{\milli\second} ~~~~~~~\\
\hline
Region of Interest Selection & $4.05$ \\
Ground Segmentation & $0.83$ \\
Background Filtering & $1.05$ \\
Outlier Removal & $4.82$ \\
Clustering & $8.00$ \\
Bounding Box Fitting & $2.15$ \\
Classification & $0.18$ \\
\hline
Total runtime & \textbf{21.08} ($47$ FPS) \\
\hline
\end{tabular}
\label{tab:results_runtime}
\end{center}
\end{table}

Table \ref{tab:results_runtime_pointpillar} shows the runtime of \textit{PointPillars} on the A9-I dataset.

\begin{table}[htbp]
\caption{Runtime evaluation of \textit{PointPillars} on the A9 intersection dataset with a single and registered point clouds. Tested in Python 3.8 and run on a NVIDIA RTX 4090.}
\begin{center}
\begin{tabular}{|l|c|c|}
\hline
\textbf{Point cloud type} ~~~~~~~~~~~~~~~& ~~~\textbf{$\varnothing$ Runtime in \si{\milli\second}} ~~~& ~~~\textbf{FPS}~~~ \\
\hline
Single LiDAR point cloud & $23.84$  & $42$\\
Registered point cloud & $26.11$  & $38$\\
\hline
\end{tabular}
\label{tab:results_runtime_pointpillar}
\end{center}
\end{table}

\begin{figure*}[t!]
\centering
\minipage{0.19\textwidth}
\caption*{a)}
  \includegraphics[width=\linewidth]{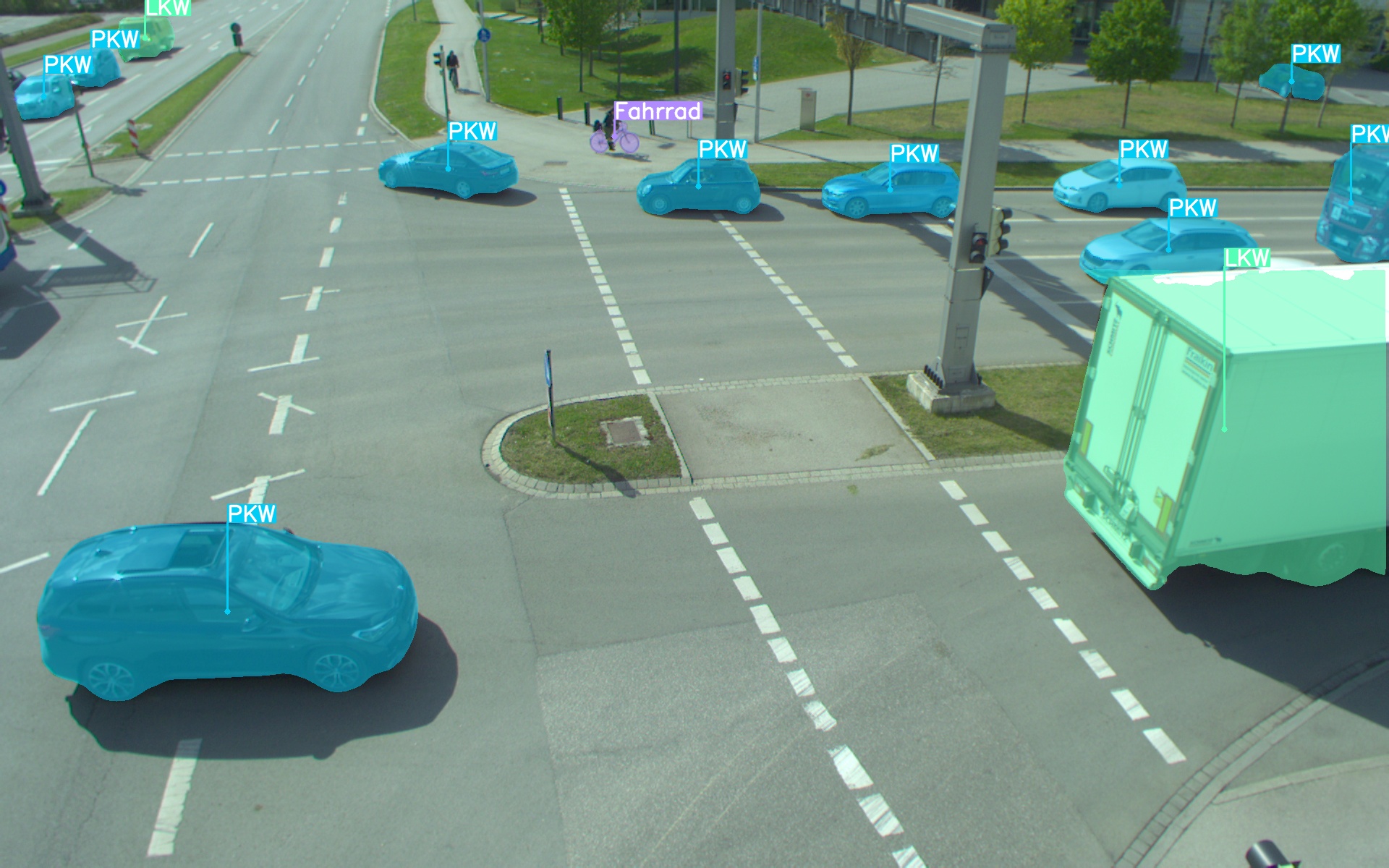}
\endminipage
\minipage{0.19\textwidth}
\caption*{b)}
  \includegraphics[width=\linewidth]{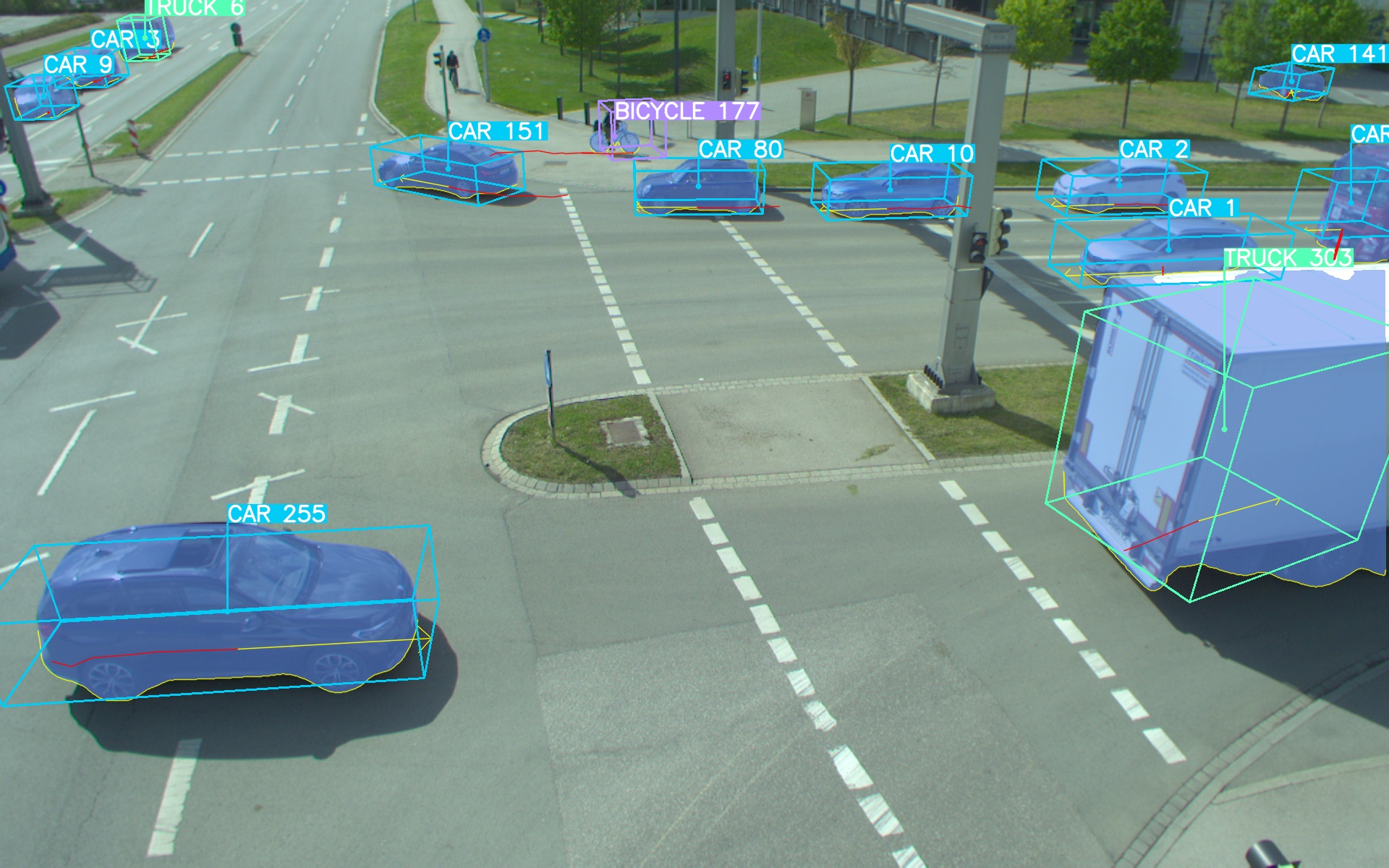}
\endminipage
\minipage{0.19\textwidth}%
\caption*{c)}
  \includegraphics[width=\linewidth]{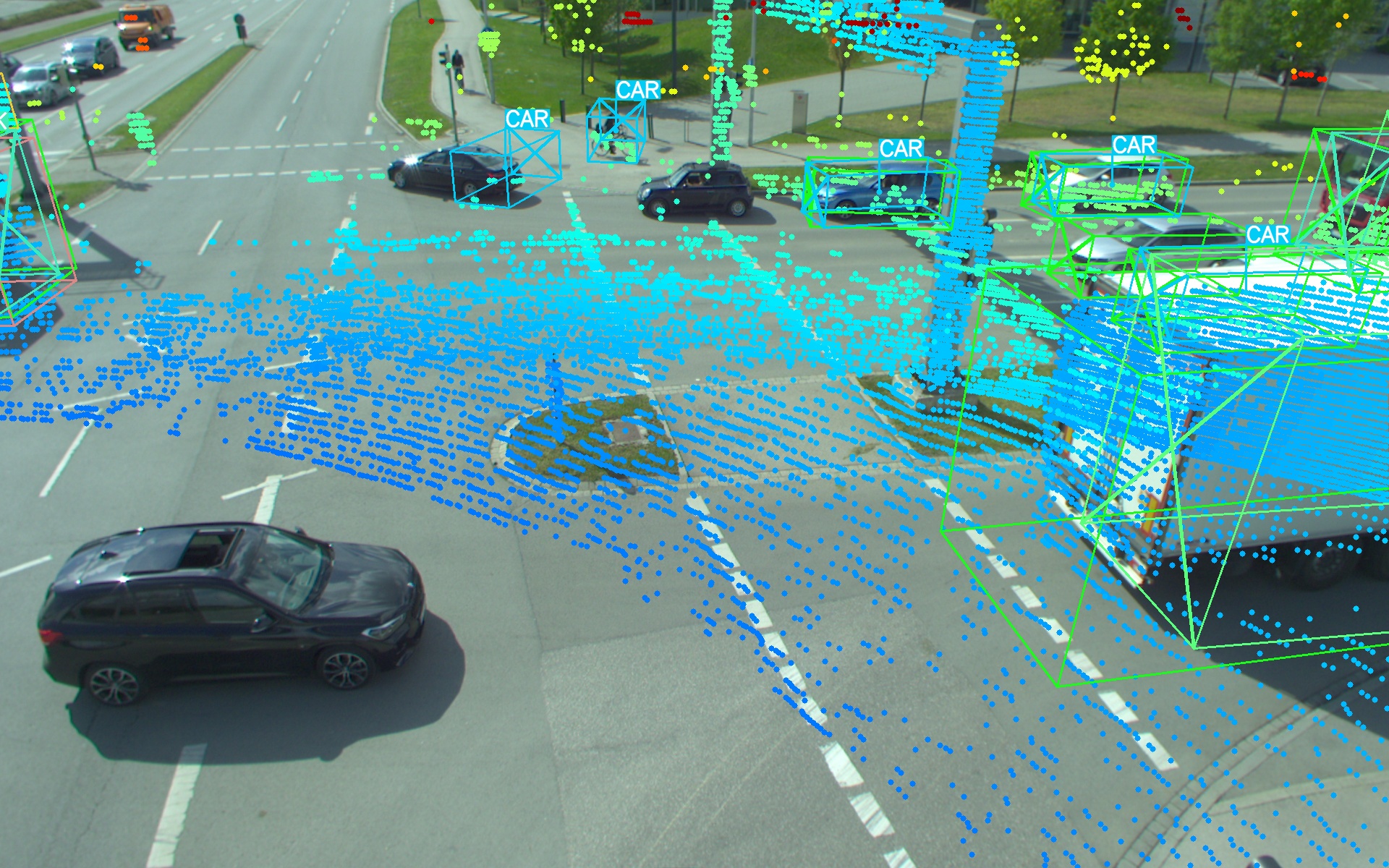}
\endminipage
\minipage{0.19\textwidth}%
\caption*{d)}
  \includegraphics[width=\linewidth]{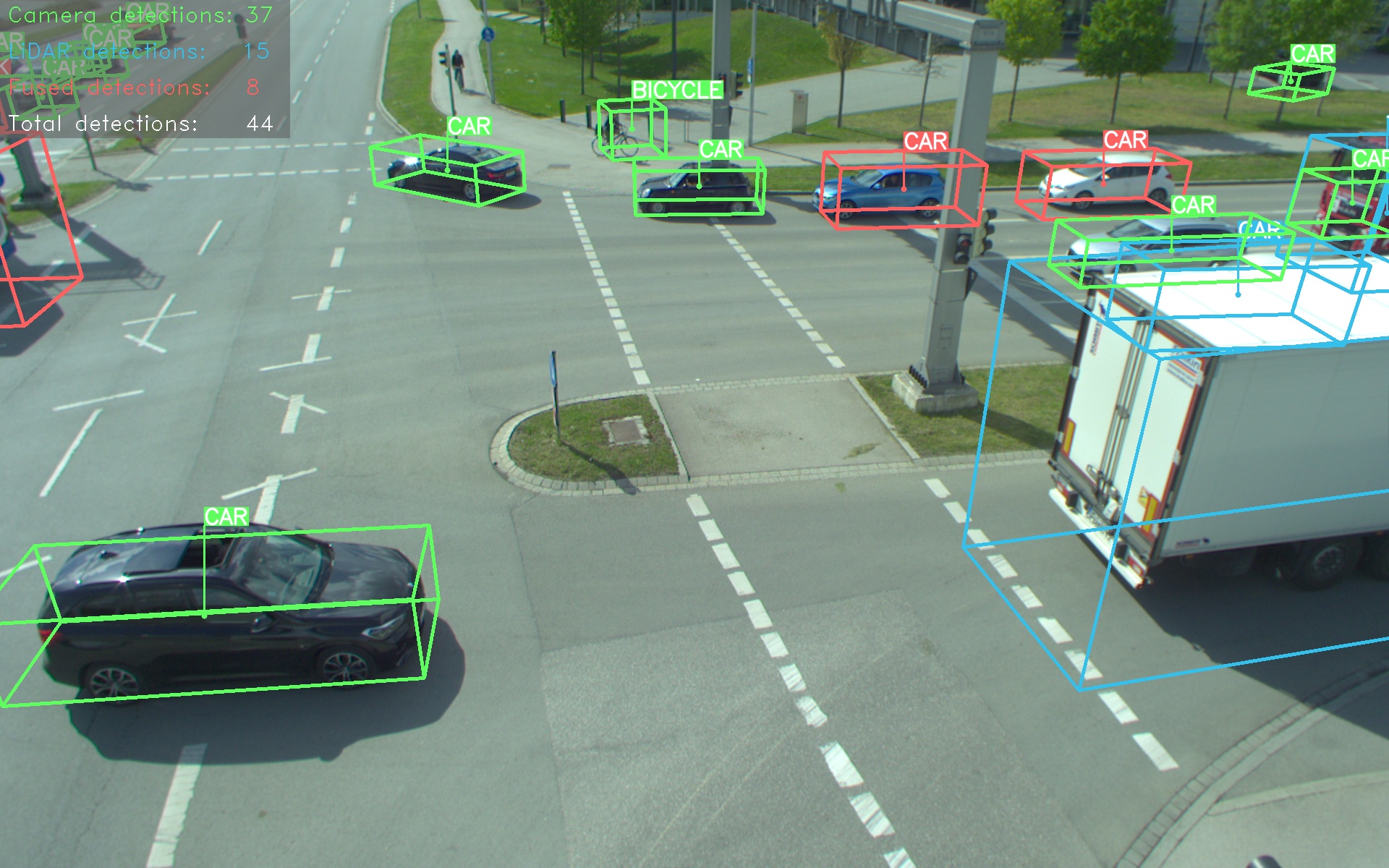}
\endminipage
\minipage{0.19\textwidth}%
\caption*{e)}
  \includegraphics[width=\linewidth]{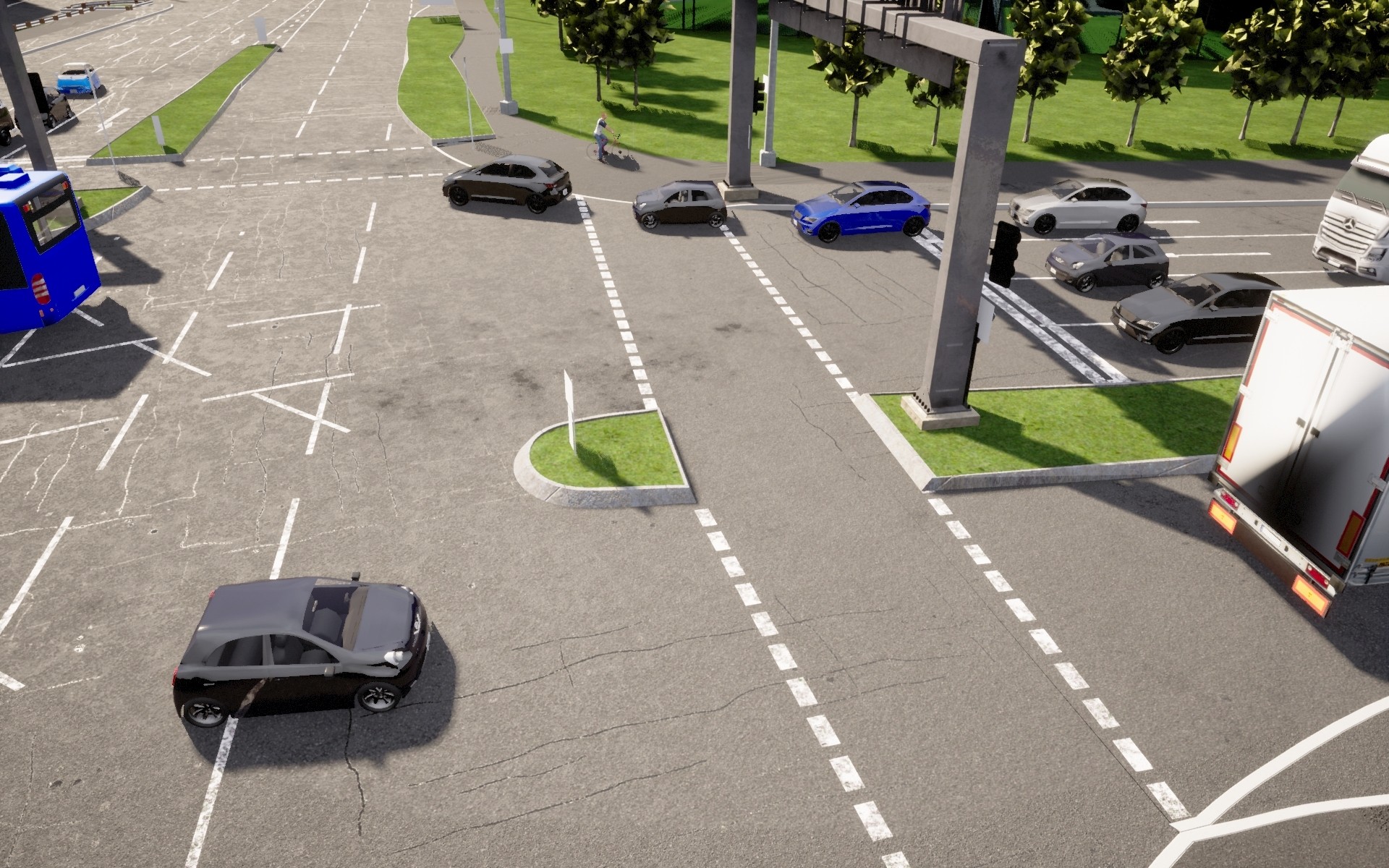}
\endminipage\hfill
\minipage{0.19\textwidth}
  \includegraphics[width=\linewidth]{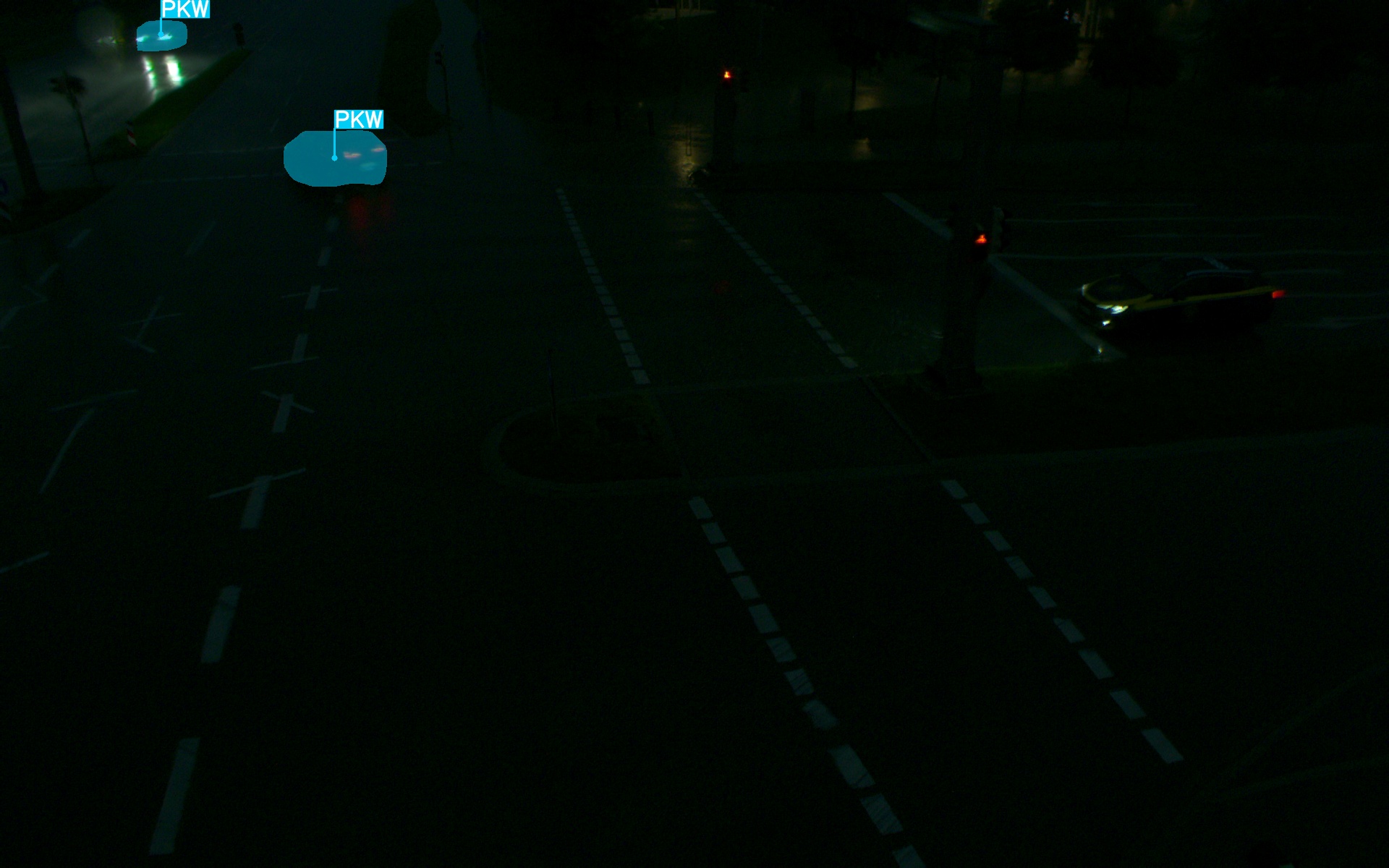}
\endminipage
\minipage{0.19\textwidth}
  \includegraphics[width=\linewidth]{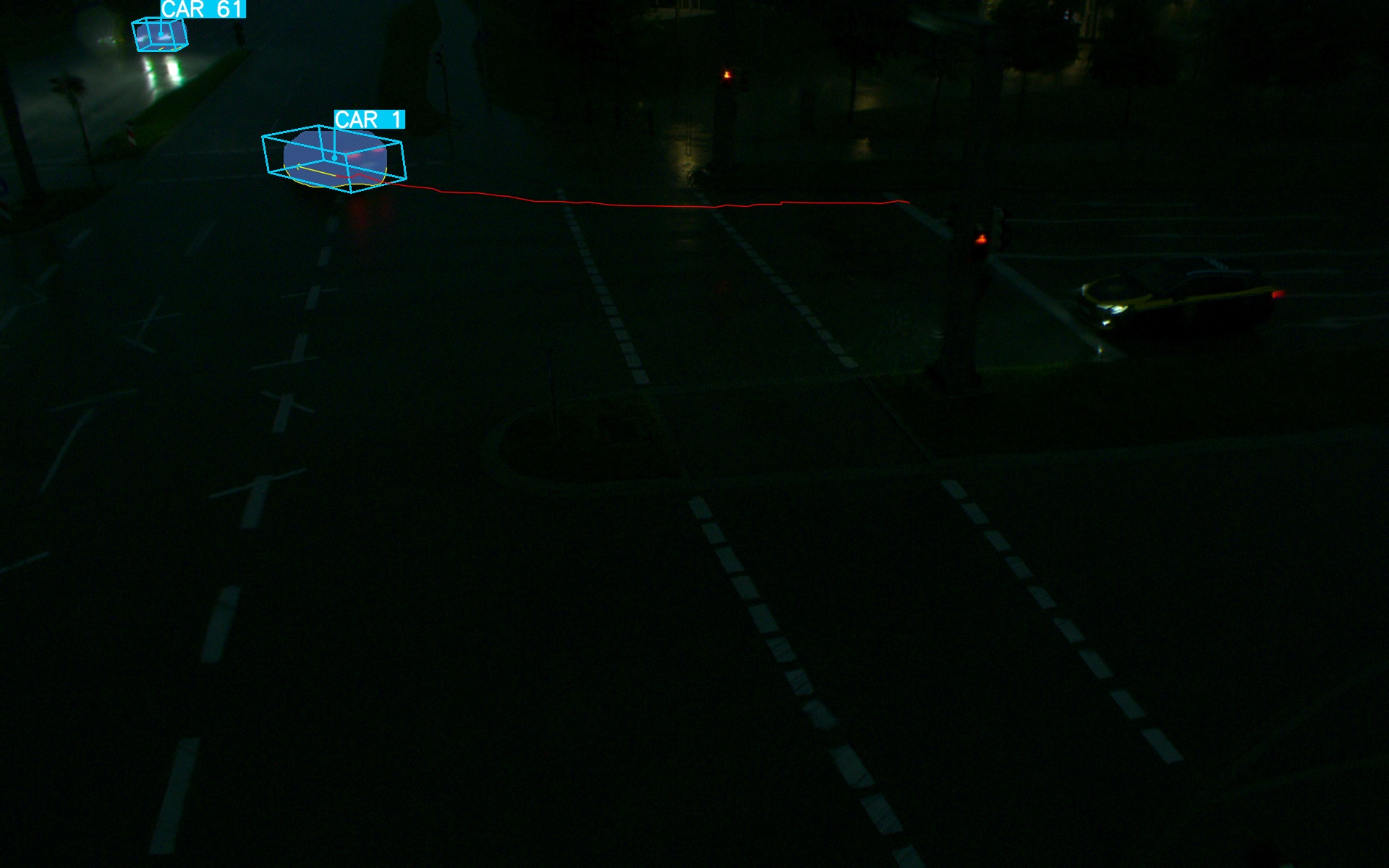}
\endminipage
\minipage{0.19\textwidth}%
  \includegraphics[width=\linewidth]{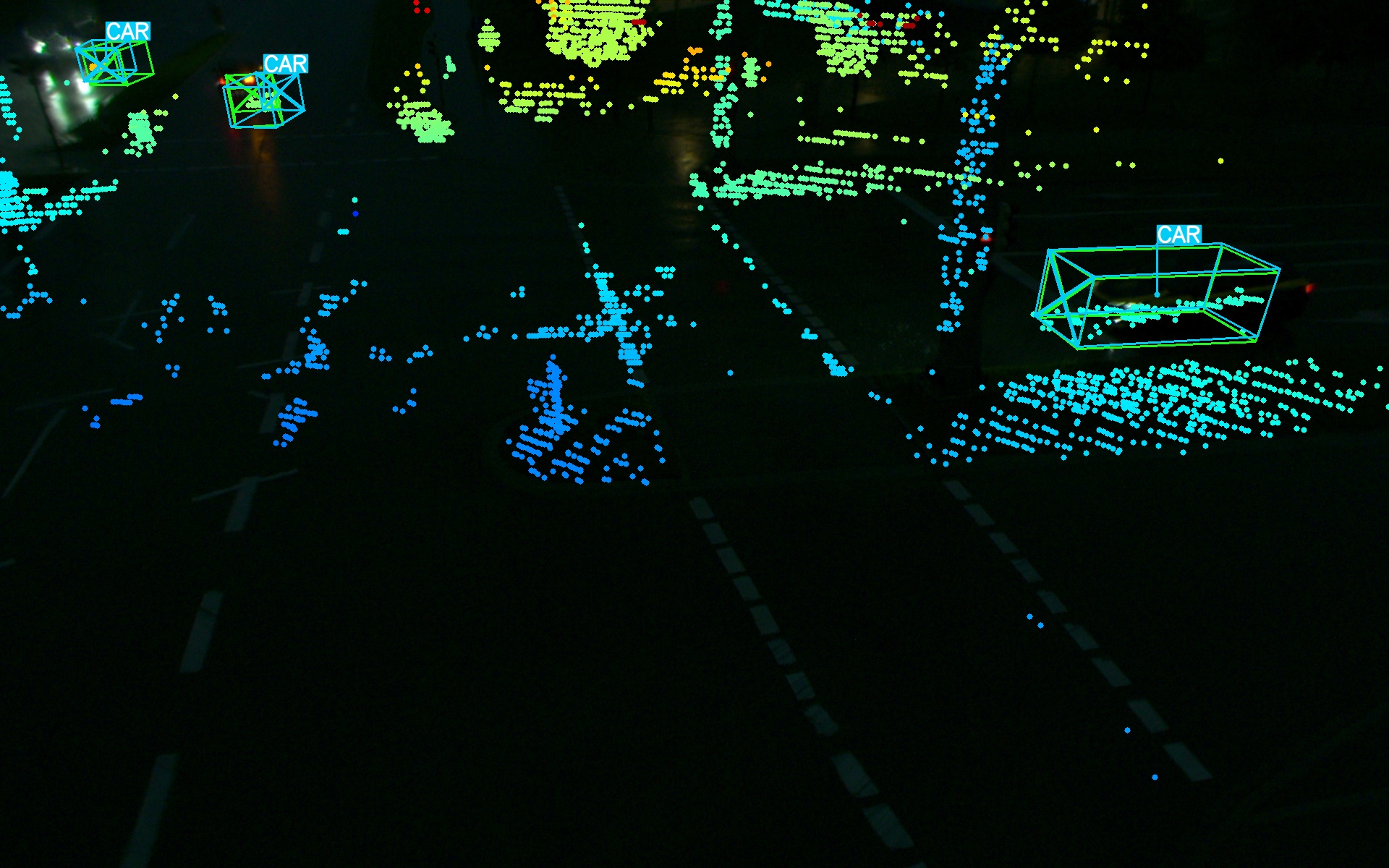}
\endminipage
\minipage{0.19\textwidth}%
  \includegraphics[width=\linewidth]{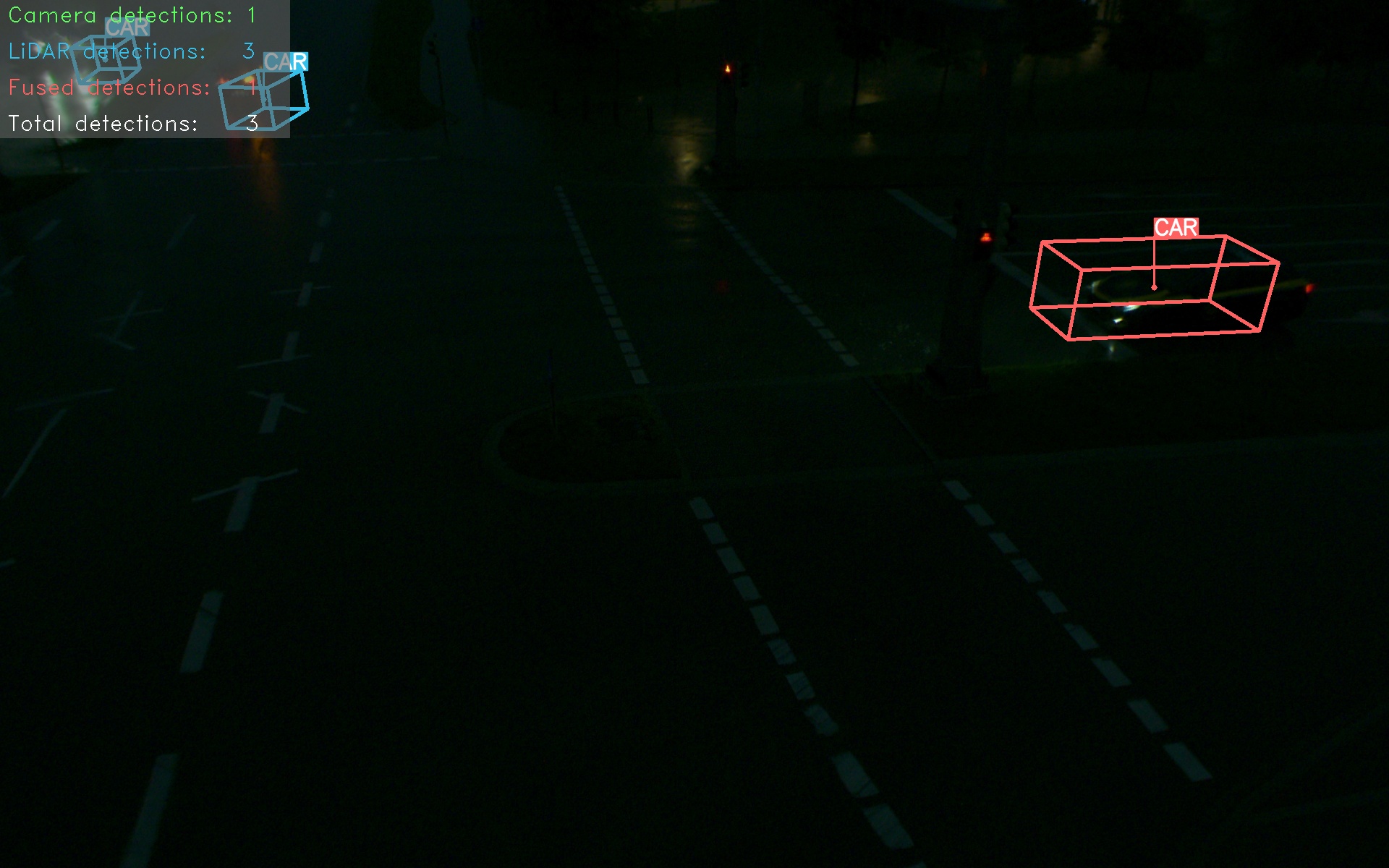}
\endminipage
\minipage{0.19\textwidth}%
  \includegraphics[width=\linewidth]{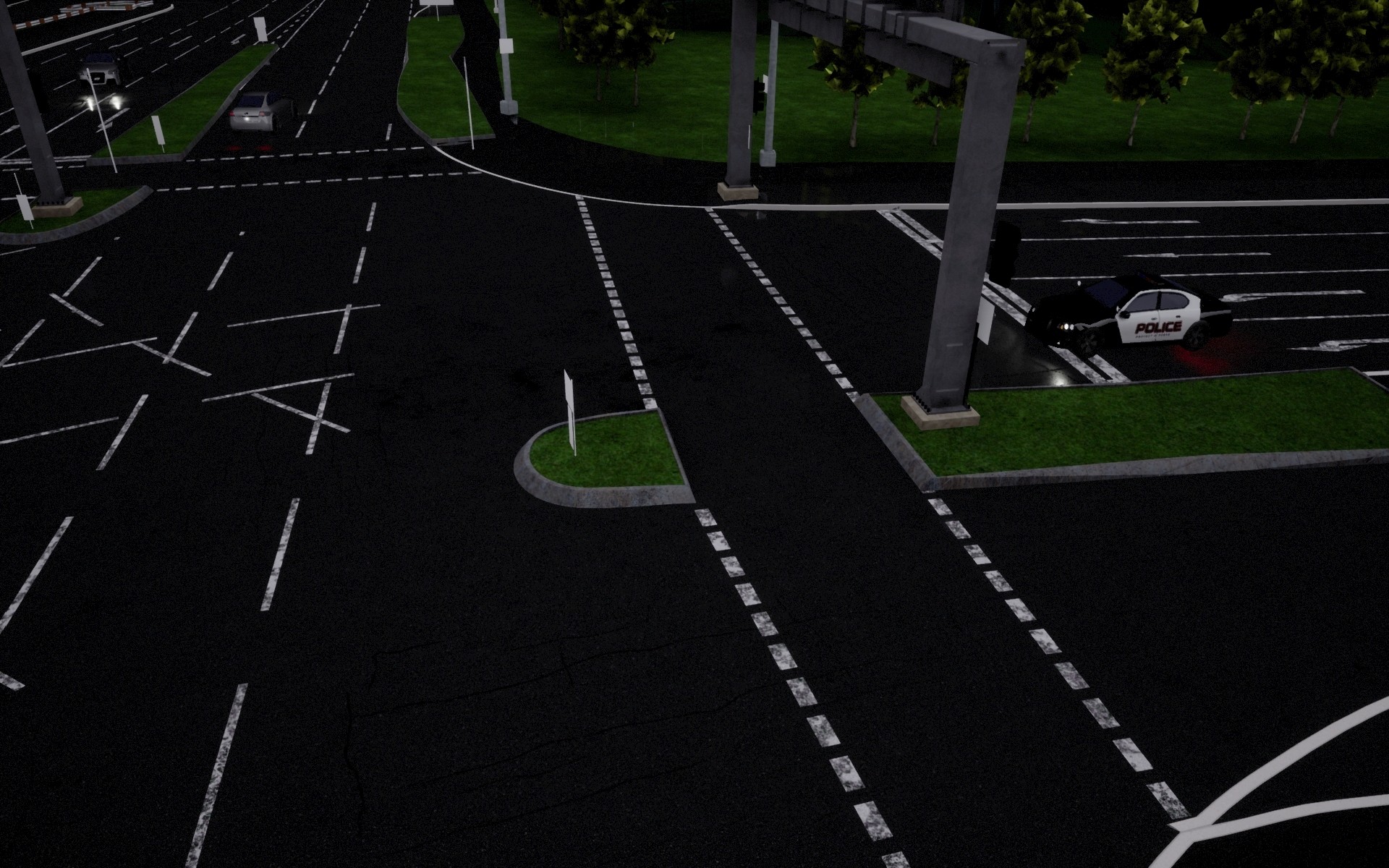}
\endminipage\hfill
\minipage{0.19\textwidth}
  \includegraphics[width=\linewidth]{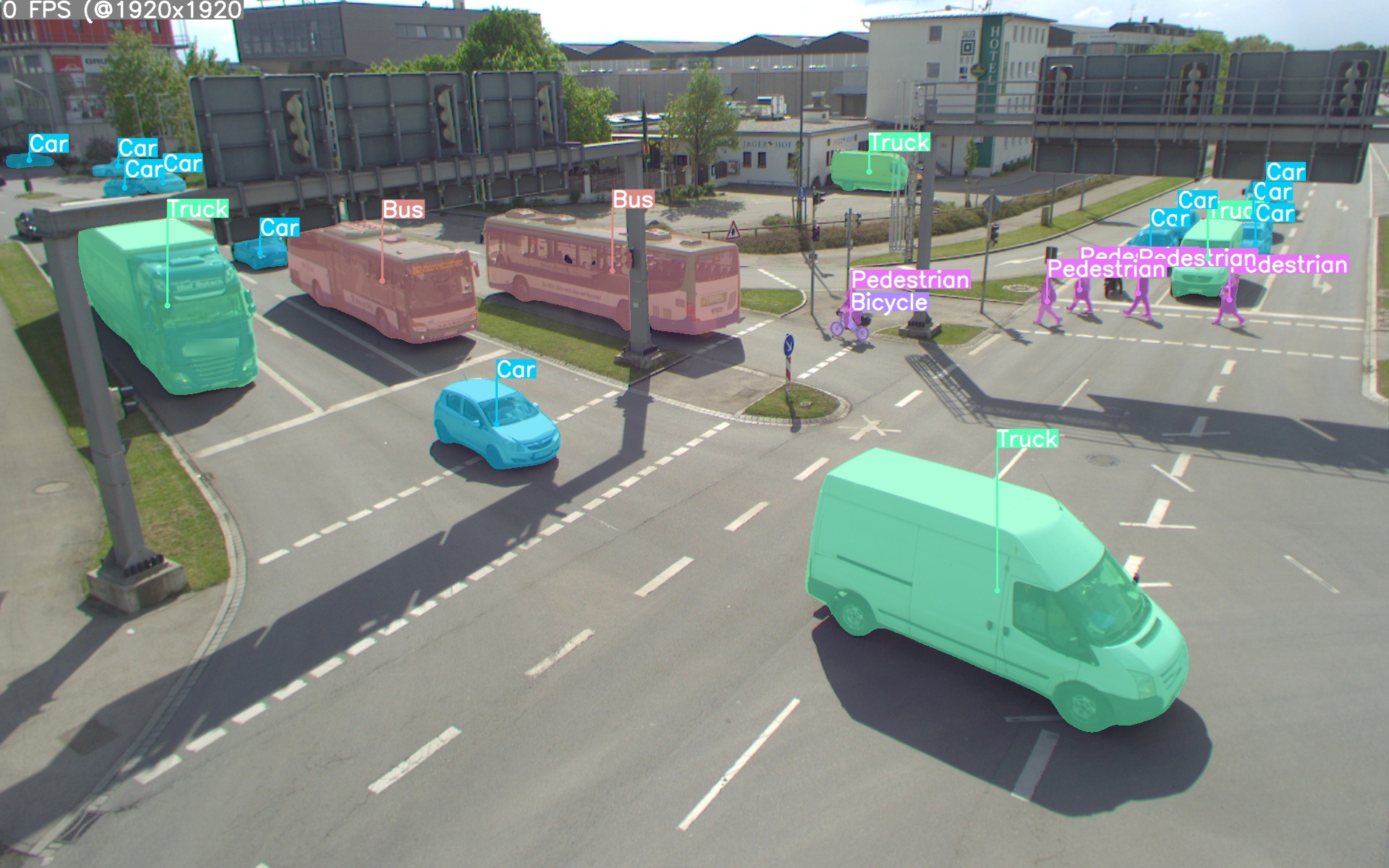}
\endminipage
\minipage{0.19\textwidth}
  \includegraphics[width=\linewidth]{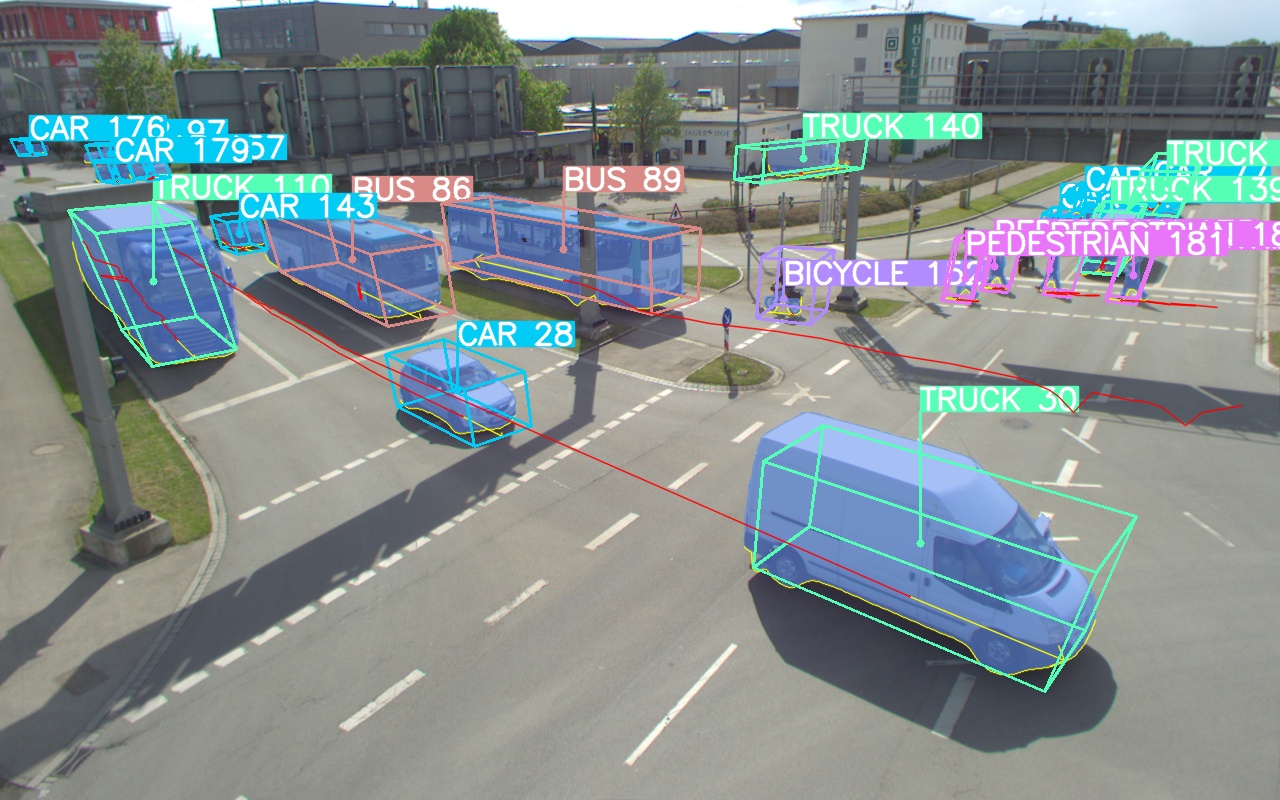}
\endminipage
\minipage{0.19\textwidth}%
  \includegraphics[width=\linewidth]{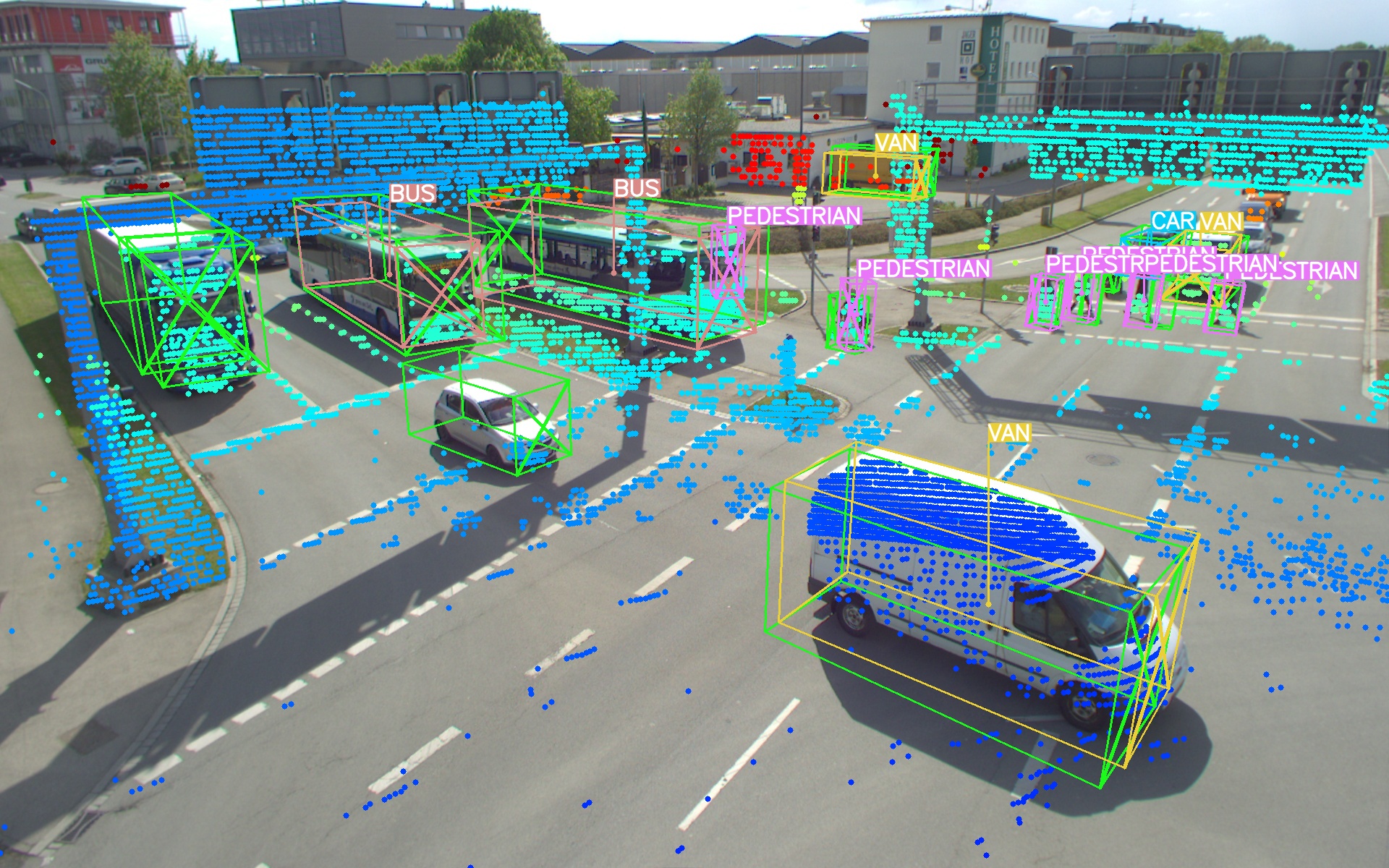}
\endminipage
\minipage{0.19\textwidth}%
  \includegraphics[width=\linewidth]{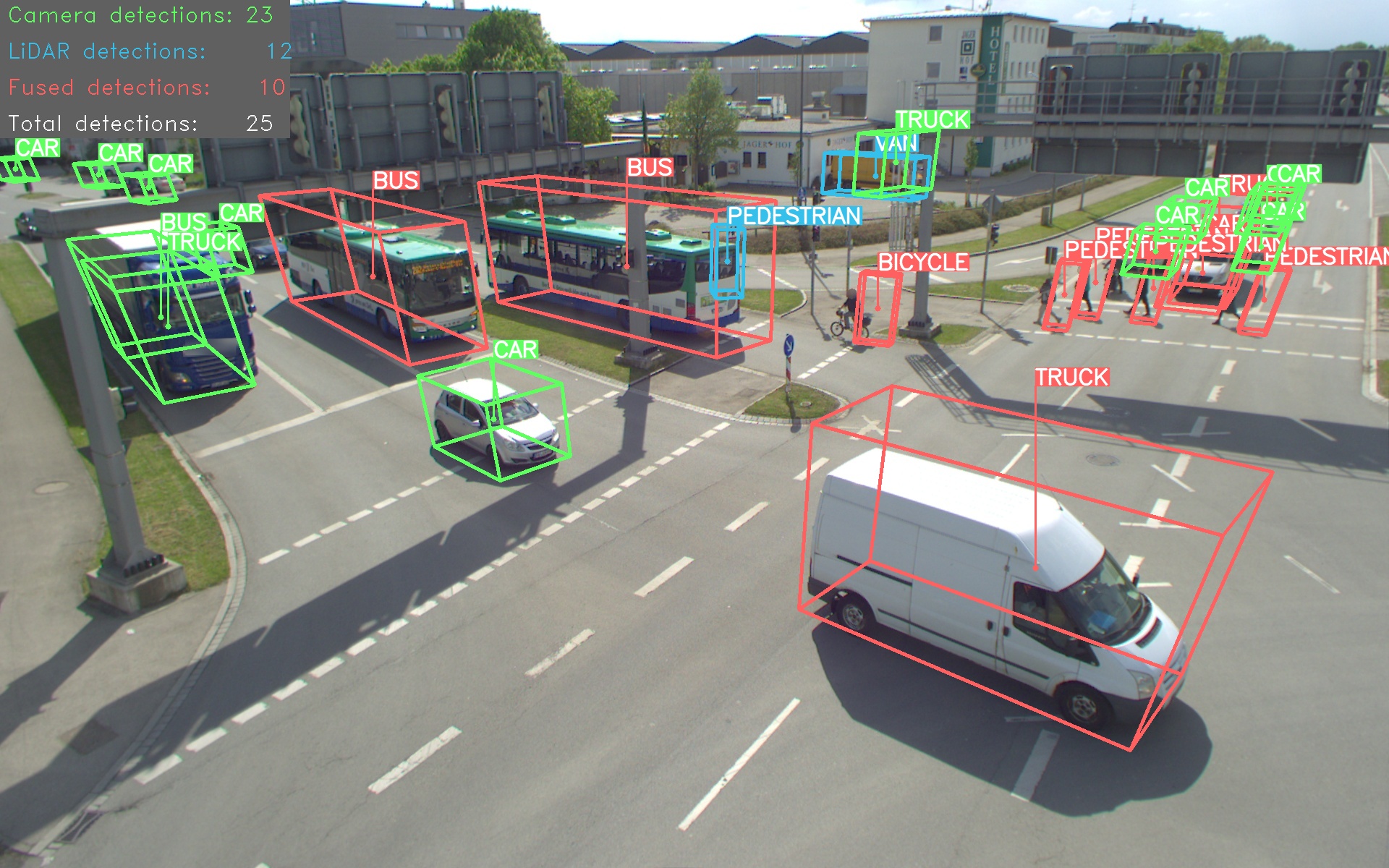}
\endminipage
\minipage{0.19\textwidth}%
  \includegraphics[width=\linewidth]{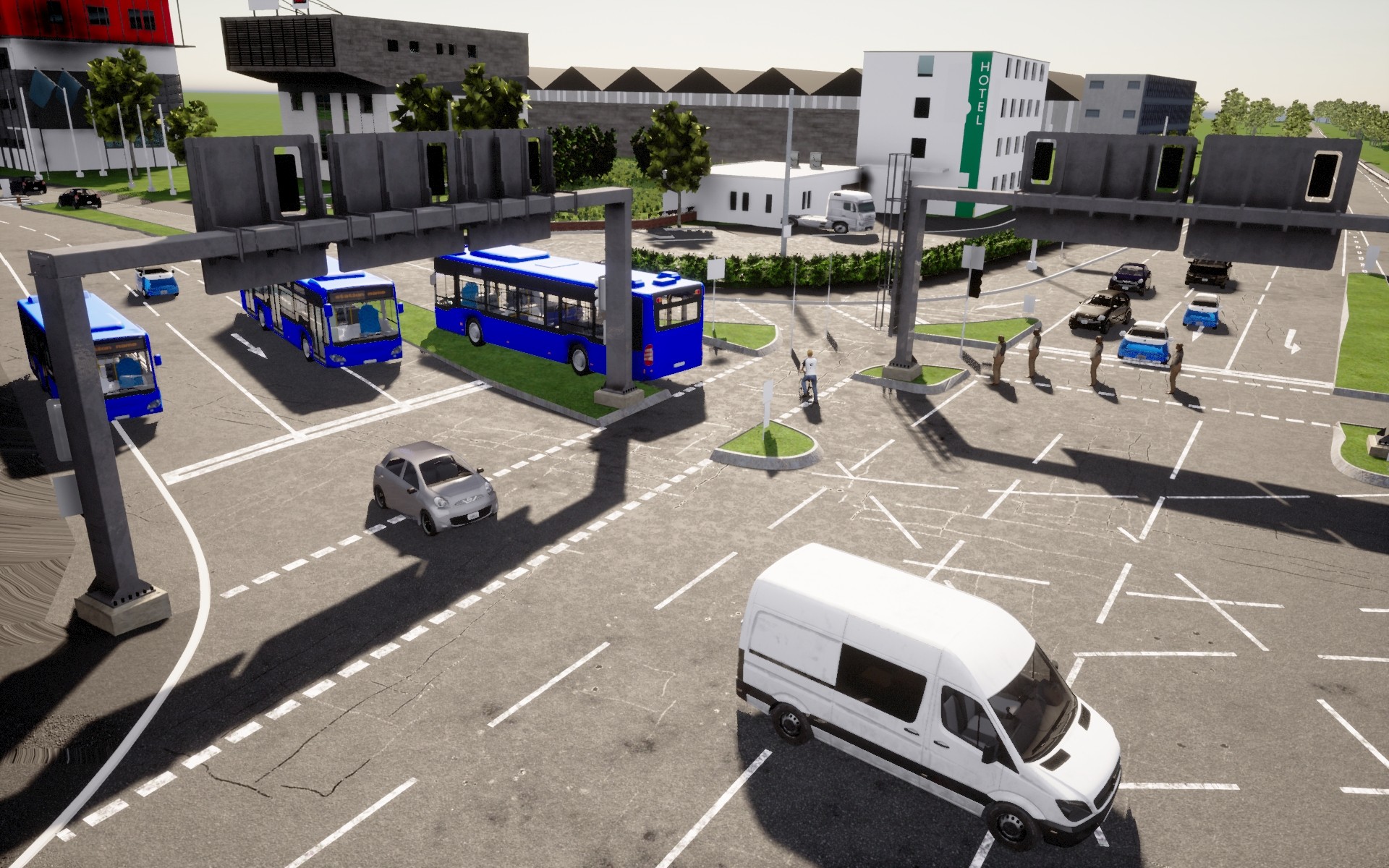}
\endminipage\hfill
\minipage{0.19\textwidth}
  \includegraphics[width=\linewidth]{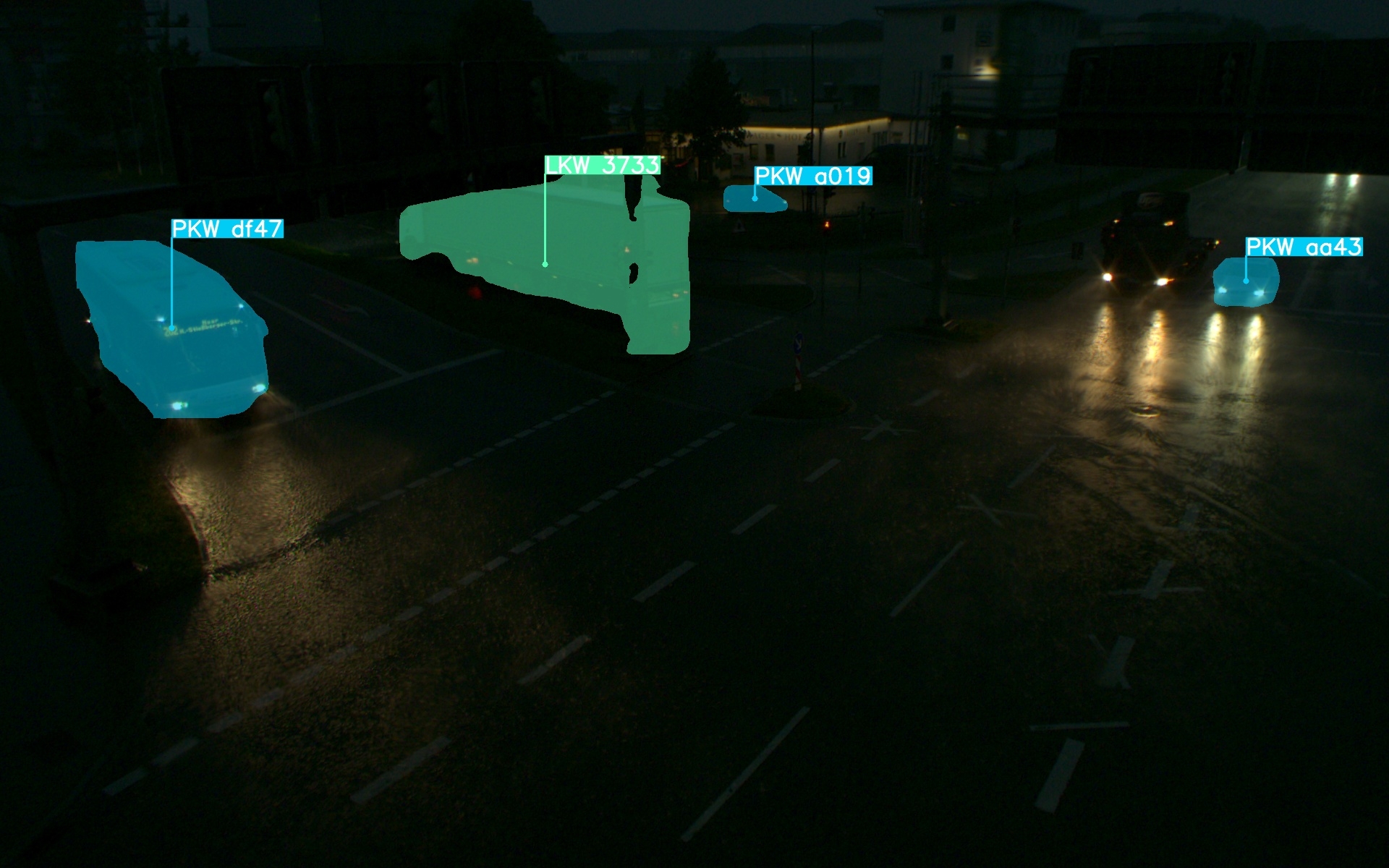}
\endminipage
\minipage{0.19\textwidth}
  \includegraphics[width=\linewidth]{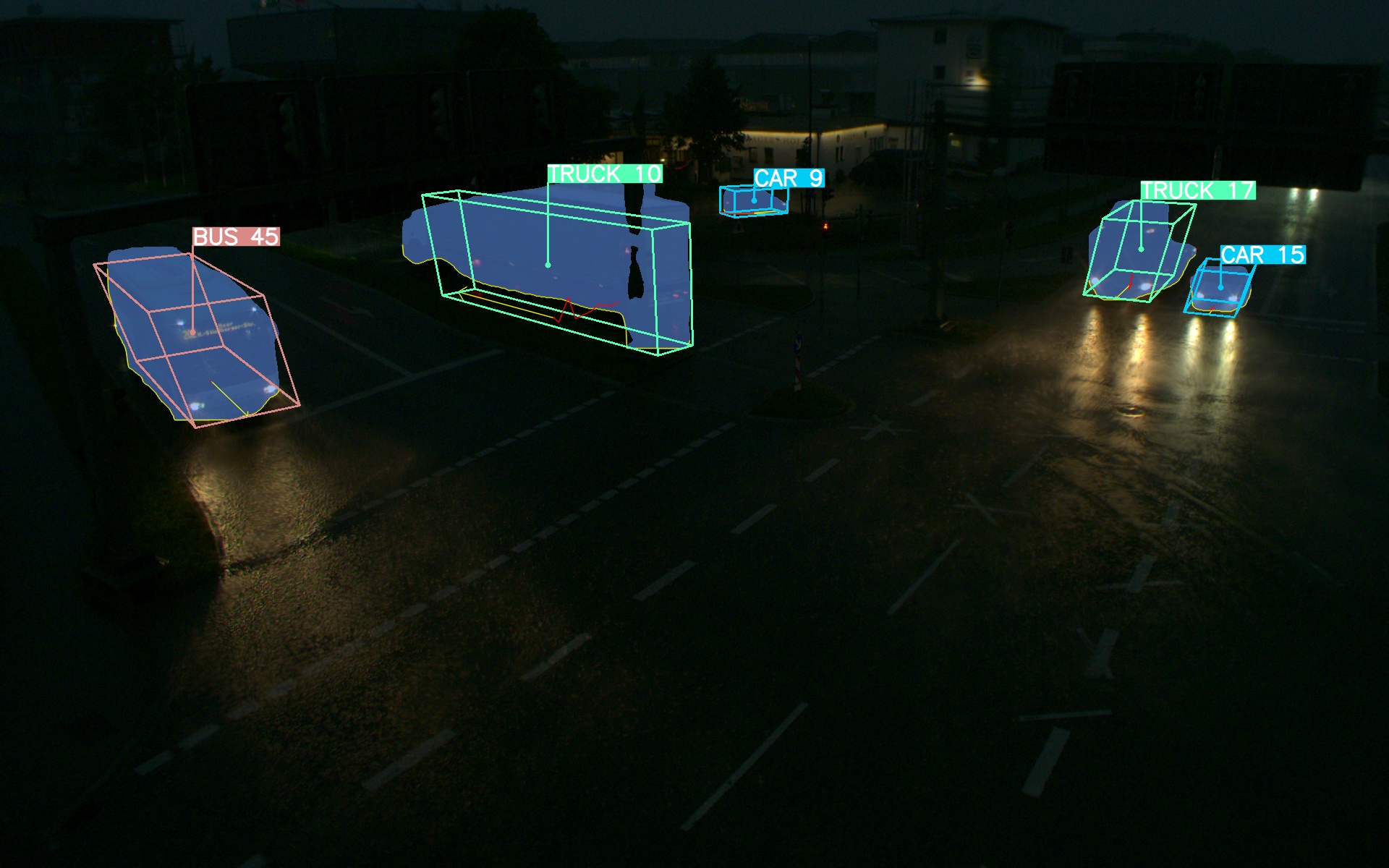}
\endminipage
\minipage{0.19\textwidth}%
  \includegraphics[width=\linewidth]{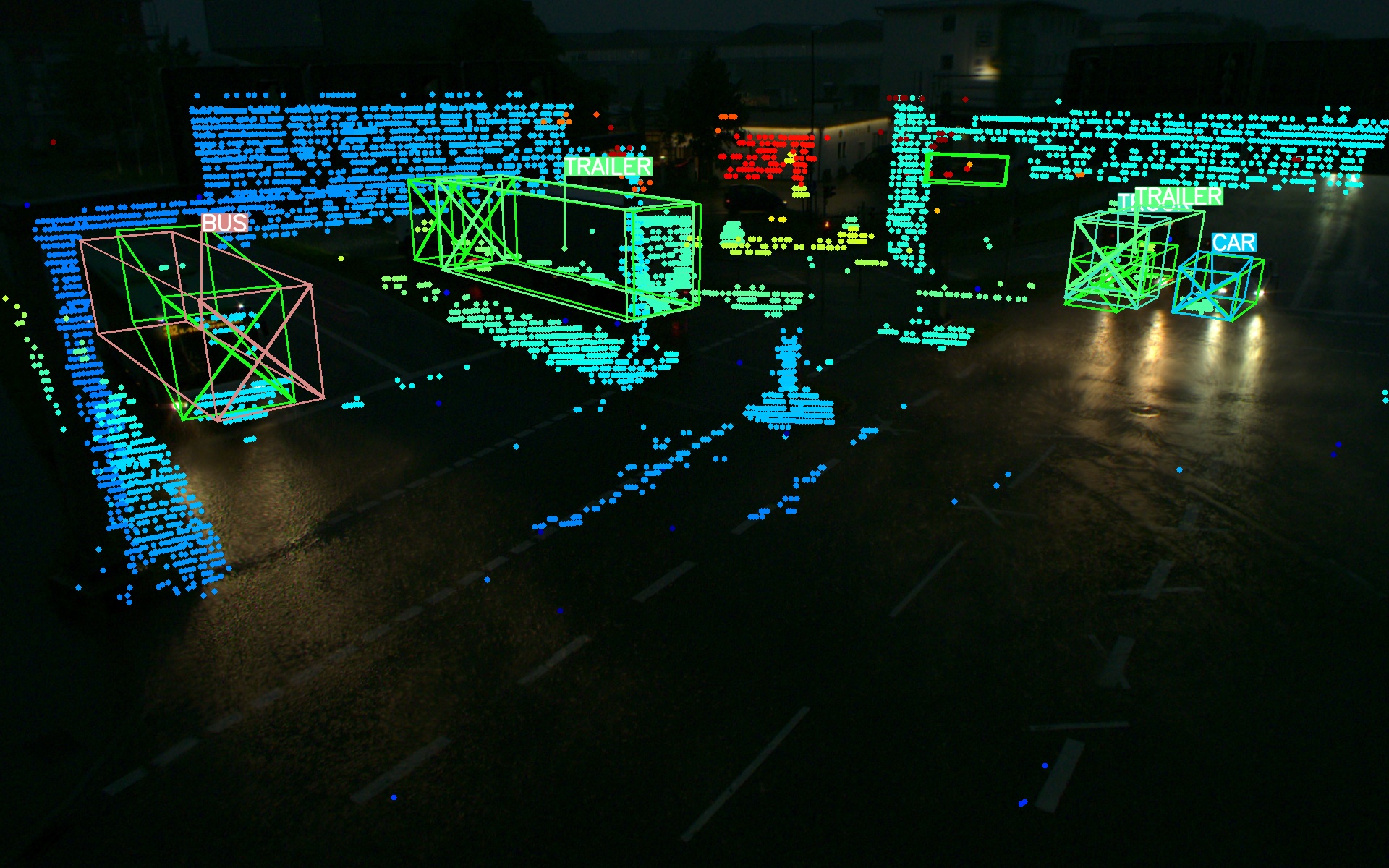}
\endminipage
\minipage{0.19\textwidth}%
  \includegraphics[width=\linewidth]{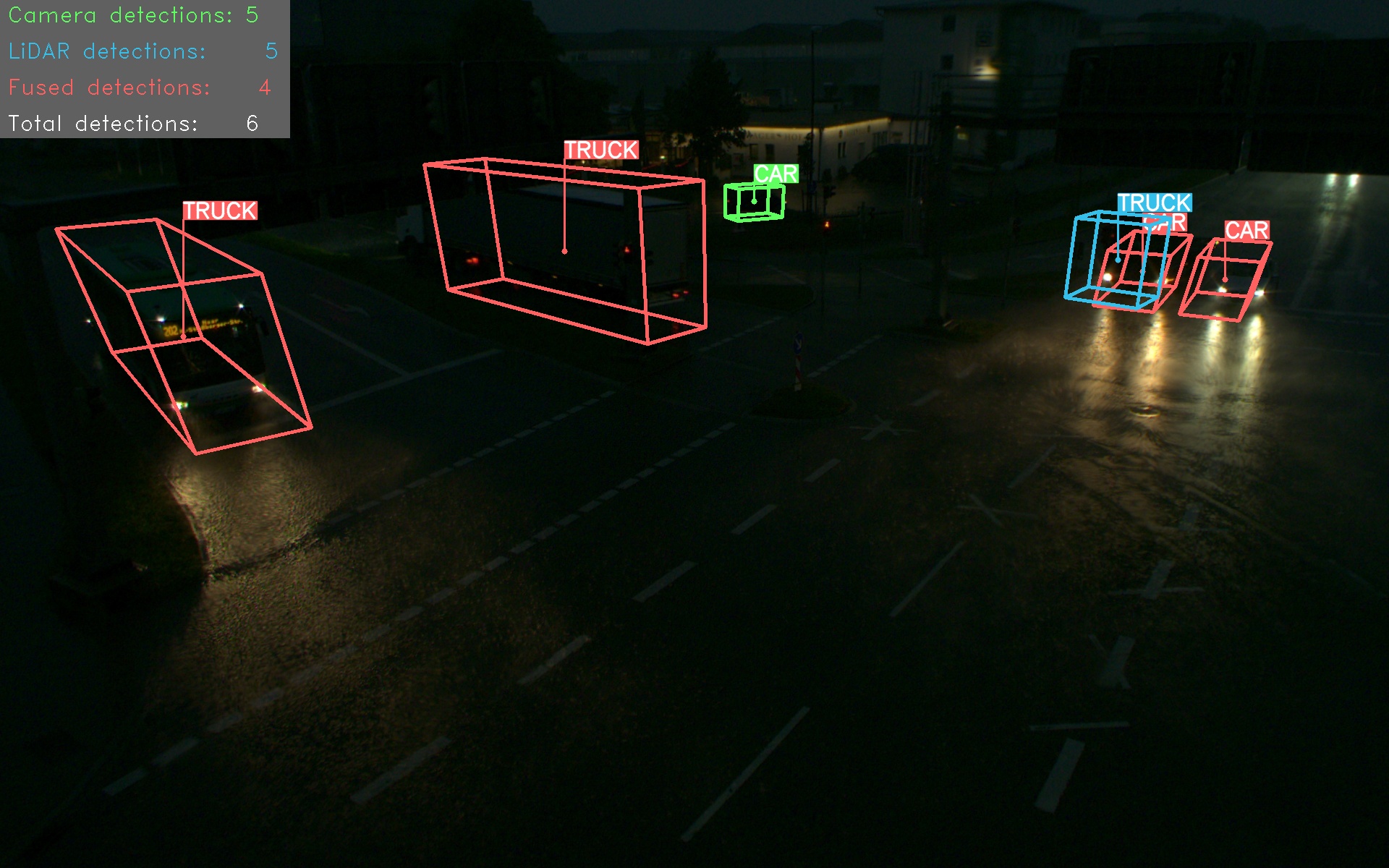}
\endminipage
\minipage{0.19\textwidth}%
  \includegraphics[width=\linewidth]{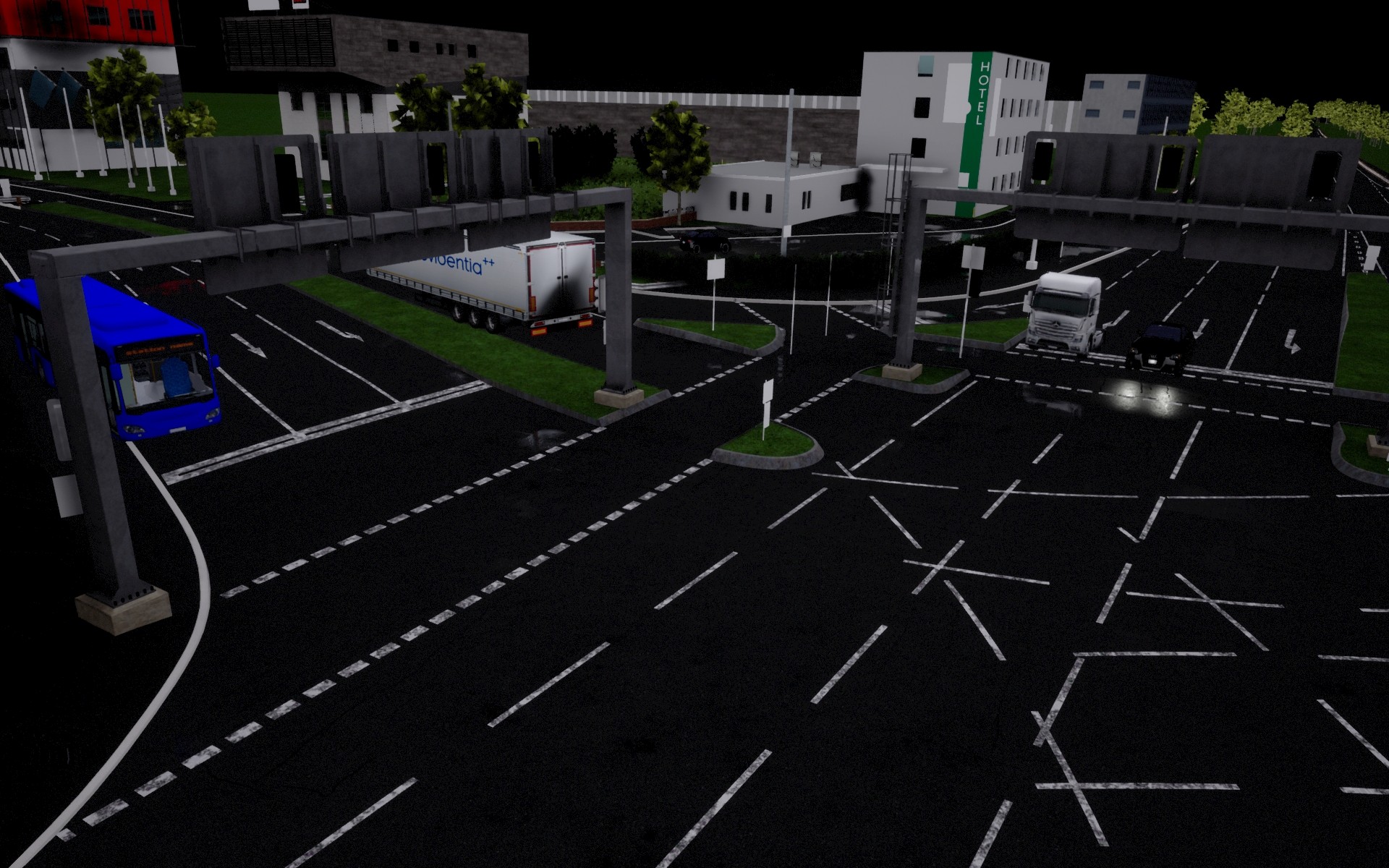}
\endminipage
   \caption{Qualitative results for south1 camera (first and second row) and south2 camera (third and fourth row).
   We show the perception results during day time (first and third row) and night time (second and fourth row). The detections are colored by their class color. Column four shows the fusion results in the following colors: green (unmatched camera detections), blue (unmatched LiDAR detections) and red (fused detections). From left to right: a) Instance segmentation, b) \textit{MonoDet3D}, c) \textit{PointPillars}, d) \textit{InfraDet3D}, e) Visualization of the fused perception results in CARLA (using early and late fusion).}
   \label{fig:qualitative_results_south1}
\end{figure*}

\subsection{Quantitative Results}
All four object detection modules were evaluated on the A9-I south1, south2 and full intersection test set (see Table \ref{tbl:evaluation}). The performance on the south1 sub set is 80\% higher on average because of better lighting conditions and 2.5x less occlusions. \textit{PointPillars} performs much better on the test set compared to \textit{LidarDet3D} since it was trained on the A9-I dataset. The south LiDAR has 2.74x more overlapping with the south2 camera which leads to higher \texttt{mAP} value ($+15.13$), compared to the north LiDAR. Using registered point clouds (early fusion) we achieve the highest results ($68.48$ \texttt{mAP}) with our \textit{InfraDet3D} fusion model on the A9-I south1 test set by fusing the camera and LiDAR detections on a late fusion level. Table \ref{tbl:class_evaluation} demonstrates that among all classes, the \textit{Bus} class exhibits the highest average precision in terms of detection accuracy since it is covered well by all LiDARs.

\begin{table}[htbp]
    \centering
    \scalebox{0.9}{
    \begin{tabular}{|l|c|c|c|c|c|c|}
        \hline
         \textbf{Class} & \textbf{Car} & \textbf{Truck} & \textbf{Motorcycle} & \textbf{Bus} & \textbf{Pedestrian} & \textbf{Bicycle} \\
         \hline
         Precision  & 71.75 & \underline{91.20} & 82.72 & \textbf{99.93} & 31.37 & 36.02 \\
         Recall  & \underline{87.33} & 85.03 & 70.71 & \textbf{100.00} & 25.49 & 80.77 \\
         \hline
         AP  & 71.64 & \underline{91.03} & 82.37 & \textbf{99.93} & 30.00 & 35.93 \\
         \hline
    \end{tabular}
    }
    \caption{Average Precision (AP) results across classes in the A9-I dataset of the best performing \textit{InfraDet3D} model.}
    \label{tbl:class_evaluation}
\end{table}

\subsection{Qualitative Results}
The qualitative results are shown in Figure \ref{fig:qualitative_results_south1}. Note that even objects outside of the sensor's field-of-view, like the black car in the first row, can be detected by fusing camera and LiDAR detections. The final perception results are visualized in the CARLA simulation environment, that contains a full reconstruction of the A9 Test Stretch.

\section{Conclusion}
\textit{InfraDet3D} is a novel perception architecture that increases situation awareness and range of traditional
single-sensor systems by combining data from multiple sensors distributed on a 20~m long infrastructure gantry bridge. We show that our multi-modal perception framework, fusing multiple roadside LiDARs and cameras, is able to achieve better results ($+1.62$ \texttt{mAP}) than object detectors using only the camera input. The distributed sensors combine their perception results and allow to detect partially and even fully occluded objects. Our solution is deployed on high performance edge units and is very cost-effective, since it is distributed among the CPU (calibration, unsupervised point cloud detection, fusion) and the GPU (instance segmentation, supervised detection in point clouds). Future trends and challenges include a better perception in adverse weather conditions such as heavy rain, snow, and fog. These conditions reduce the range, reflection intensity, resolution of point clouds, increase the noise, and produce outliers. In \cite{le2022adaptive} and \cite{park2020fast}, a method to filter snow points is proposed that will be incorporated in the future. A point cloud compression module will be integrated for real-time communication and data sharing between RSUs and vehicles. In the future, we plan to extend our framework into a deep fusion architecture. Finally, our goal is to evaluate our models on other infrastructure roadside datasets like DAIR-V2X-I \cite{yu2022dair}, Rope3D \cite{ye2022rope3d}, LUMPI \cite{busch2022lumpi}, and IPS300+ \cite{wang2022ips300}. We will also label more roadside sensor data and apply few-shot and active learning \cite{hekimoglu2022efficient} to deal with small datasets and limited information. To improve domain adaptation, we will adapt our solution to other roadside LiDAR sensors and different domains (ODDs) to achieve a domain-invariant data representation.

\section*{Acknowledgment}
This research was supported by the Federal Ministry of Education and Research in Germany within the project \textit{AUTOtech.agil}, Grant Number: 01IS22088U. We thank Christian Creß, Venkatnarayanan Lakshminarasimhan, and Leah Strand for the collective work on the A9 infrastructure system. Moreover, we thank 3D Mapping Solutions for providing the HD map.

\addtolength{\textheight}{-2.5cm}



\end{document}